\documentclass[journal]{IEEEtran}

\usepackage{cite}
\usepackage{graphicx}
\usepackage{textcomp}
\usepackage{wrapfig}
\usepackage[utf8]{inputenc}
\usepackage{authblk}
\usepackage{graphicx}
\usepackage{cite}
\usepackage{indentfirst}
\usepackage[justification=centering]{caption}
\usepackage{amsmath}
\usepackage{lscape}
\usepackage{gensymb}
\usepackage{algorithm}
\usepackage{algpseudocode}
\usepackage{mathtools}
\usepackage{amsfonts}
\usepackage{comment}
\usepackage{bm}
\usepackage{multirow}
\usepackage{marvosym}
\usepackage[colorlinks=true, linkcolor=black, citecolor=black, urlcolor=black]{hyperref}
\usepackage[dvipsnames]{xcolor}
\algnewcommand{\Inputs}[1]{%
  \State \textbf{Inputs:}
  \Statex \hspace*{\algorithmicindent}\parbox[t]{
.8\linewidth}{\raggedright #1}
}
\algnewcommand{\Initialize}[1]{%
  \State \textbf{Initialize:}
  \Statex \hspace*{\algorithmicindent}\parbox[t]{.8\linewidth}{\raggedright #1}
}
\usepackage[dvipsnames]{xcolor}
\DeclareMathOperator*{\argmin}{arg\,min}  
\usepackage{eso-pic}

\newcommand\WatermarkText{This paper has been accepted for publication at IEEE Transactions on Image Processing (TIP), 2024. ©IEEE}

\AddToShipoutPictureBG*{%
  \AtPageUpperLeft{%
    \setlength{\unitlength}{1cm}
    \put(0, -1.5){\parbox{\paperwidth}{\centering \textcolor{gray}{%
      \begin{minipage}{0.9\textwidth}
        \centering
        \WatermarkText
      \end{minipage}%
    }}}
  }
}

\graphicspath{ {Images/} }
\graphicspath{ {Images/} }

\begin{document}

\title{E-Calib: A Fast, Robust and Accurate Calibration Toolbox for Event Cameras}

\author{Mohammed Salah,
        Abdulla Ayyad,
        Muhammad Humais,
        Daniel Gehrig,
        Abdelqader Abusafieh,
        Lakmal~Seneviratne,
        Davide Scaramuzza,
        and~Yahya~Zweiri
\thanks{This work was supported by by the Advanced Research and Innovation Center (ARIC), which is jointly funded by STRATA Manufacturing PJSC (a Mubadala company), Khalifa University of Science and Technology in part by Khalifa University Center for Autonomous Robotic Systems (KUCARS) under Award RC1-2018-KUCARS, and Sandooq Al Watan under Grant SWARD-S22-015.}
\thanks{M. Salah, A. Ayyad, and Y. Zweiri are with ARIC, Khalifa University, Abu Dhabi, UAE. Y. Zweiri is also with the Department of Aerospace Engineering and is the director of ARIC.

M. Humais and L. Seneviratne are with KUCARS, Khalifa University, Abu Dhabi, UAE, and L. Seneviratne is the director of KUCARS.

 D. Gehrig and D. Scaramuzza are with the Robotics and Perception Group, University of Zurich, Switzerland.
 
 A. Abusafieh is with Research and Development, Strata Manufacturing PJSC (a Mubadala company), Al Ain, UAE.
 
 Mohammed Salah is the corresponding author (email: mohammed.salah@ku.ac.ae)}}

\maketitle

\begin{abstract}
Event cameras triggered a paradigm shift in the computer vision community delineated by their asynchronous nature, low latency, and high dynamic range. Calibration of event cameras is always essential to account for the sensor intrinsic parameters and for 3D perception. However, conventional image-based calibration techniques are not applicable due to the asynchronous, binary output of the sensor. The current standard for calibrating event cameras relies on either blinking patterns or event-based image reconstruction algorithms. These approaches are difficult to deploy in factory settings and are affected by noise and artifacts degrading the calibration performance. To bridge these limitations, we present E-Calib, a novel, fast, robust, and accurate calibration toolbox for event cameras utilizing the asymmetric circle grid, for its robustness to out-of-focus scenes. E-Calib introduces an efficient reweighted least squares (eRWLS) method for feature extraction of the calibration pattern circles with sub-pixel accuracy and robustness to noise. In addition, a modified hierarchical clustering algorithm is devised to detect the calibration grid apart from the background clutter. The proposed method is tested in a variety of rigorous experiments for different event camera models, on circle grids with different geometric properties, on varying calibration trajectories and speeds, and under challenging illumination conditions. The results show that our approach outperforms the state-of-the-art in detection success rate, reprojection error, and pose estimation accuracy.
\end{abstract}

\IEEEpeerreviewmaketitle

\section*{Code and Multimedia Material}

\begin{center}

Video:

\url{https://youtu.be/4giQn6rt-48}

Code and Dataset:

\url{https://github.com/mohammedsalah98/E_Calib}

\end{center}

\section{Introduction} \label{sec:introduction}
\IEEEPARstart {O}{ver} the past decades, imaging sensors have evolved at a rapid pace providing solutions for various intelligent perception algorithms. The event camera is a bio-inspired sensor whose pixels asynchronously fire events upon change of light intensity. Each generated event records the coordinates of the pixel spiking, the time at which the event occurred, and the sign of the intensity change. The output of an event camera is therefore not frames at constant time intervals but rather an asynchronous stream of events in space and time. Due its asynchronous and differential nature, an event camera has three key advantages for robot manufacturing: sub-millisecond latency, microsecond resolution, and very high dynamic range \cite{event_survey, vision_transformer, tobi_dvs}. These features grant the sensor robustness against motion blur and poorly lighted environments, marking a new paradigm for vision sensors. With such unprecedented capabilities in hand, event-based vision unveiled its impact in space robotics \cite{nvbm_tim, scaramuzza_space, neutron_nvs}, robotic manufacturing \cite{ayyad_drilling}, fast object detection and tracking \cite{tip_detection, tip_tracking, rvt, scar_tracking}, autonomous navigation \cite{event_odometry_1, scar_auto2, scar_autonomous}, and indoor positioning systems \cite{flicker2}.

In all of the aforementioned applications, calibration of the event camera is a necessary prerequisite to account for sensor intrinsics and for pose estimation. Due to the asynchronous, spatially sparse, binary output of events, conventional calibration methods are not applicable for calibrating event cameras. Calibration of the earliest event camera models was attained using blinking calibration patterns \cite{ecalib_1}, but require extensive instrumentation. A more practical method evolved relying on event-based image reconstruction algorithms \cite{eventcalib_e2vid}, where calibration was attained using conventional image-based calibration methods. Nevertheless, the reconstructed images suffer from artifacts and sensor noise degrading the calibration accuracy.

More sophisticated event camera models, also called Dynamic Active-Pixel Vision Sensor (DAVIS) \cite{davis}, are accompanied by frame-based imaging capabilities, where the active-pixel sensor (APS) frames are utilized for calibration using conventional calibration techniques. As event-based vision is gradually gaining maturity, recent models of event cameras are rather manufactured as stand-alone sensors. Thus, Huang et al. \cite{ecalib_iros} proposed a multi-segment optimization method to obtain the event camera intrinsics from the raw events using the asymmetric circle grid for its robustness against defocused scenes. However, the approach utilizes the fired raw events for the calibration optimization. Hence, sub-pixel localization accuracy of the calibration feature points is not attained degrading the calibration accuracy. This ultimately instigates a surging demand for a robust, event-driven calibration tool to obtain the intrinsic parameters of event cameras.

The drawbacks of the previously mentioned methods show that an accurate calibration tool is needed for event cameras and is attained with the following: 1) Robustness to sensor noise and 2) sub-pixel localization of the calibration pattern feature points. Accordingly, this paper presents E-Calib with these attributes and the following contributions:

\begin{enumerate}
    \item An efficient reweighted least squares (eRWLS) method is proposed to extract the feature points of the calibration targets in spatiotemporal domain with sub-pixel localization accuracy and robustness to noise. 
    \item A modified hierarchical clustering algorithm is devised to detect the calibration grid apart from the presence of outliers and background clutter.
        \item We validated the proposed work in rigorous experiments on two event camera models, using different calibration patterns, under challenging lighting conditions, and on three calibration trajectories at different speeds. The results show that our approach outperforms the state-of-the-art in terms of detection success rate, reprojection error, extrinsics estimation accuracy, and computational efficiency.
    \item We release the source code of E-Calib for the community and provide an easy to use calibration toolbox for event cameras. In addition, we release \textit{ECam\_ACircles}, a calibration dataset featuring calibration sequences with varying resolution, circle patterns, and lighting conditions with the ground truth pose of the camera for benchmarking of future works both available at: \url{https://github.com/mohammedsalah98/E_Calib}.
\end{enumerate}

\subsection{Related Work} \label{subsection:related_work}
Camera calibration recently has matured in the computer vision community with numerous algorithms have been developed to obtain the intrinsic parameters for imaging sensors \cite{cam_calib_1, cam_calib_2, person}. Intrinsic calibration is acquired with a camera observing anchor points on a calibration pattern, where the checkerboard \cite{checkerboard_calib}, coplanar circles \cite{circles_calib, tip_spheres}, and AprilTags \cite{arpiltag_calib} grids are usually utilized. For instance, conventional camera calibration became trivial and current libraries such as OpenCV \cite{calib_opencv} and MATLAB \cite{calib_matlab} provide open source calibration toolboxes.

Although image processing techniques have advanced in the past decades, the unconventional nature of event cameras inhibits the applicability of such algorithms for event-based vision. Still, knowledge of the sensor intrinsic parameters remains necessary for developing robust event-based perception algorithms. The first method developed for calibrating event cameras utilized blinking calibration patterns \cite{ecalib_1}. Since event-based vision sensors require light intensity variation to generate events, blinking screens alleviate this restriction by consistently firing events resembling checkerboard and circle calibration grids. Nonetheless, such approach needs unnecessary instrumentation as specialized screens are required to render the calibration pattern at specific frequencies \cite{eventcalib_e2vid}. More importantly, it is usually desired to move the calibration pattern instead of the event camera during calibration. This is the case when the imaging sensor is mounted on unmanned aerial and ground vehicles \cite{move_pattern}, where manipulating the blinking screen is not feasible. Due to the limitations of blinking calibration patterns, recent advances in deep learning have been utilized to reconstruct grayscale frames, where conventional calibration methods are utilized to obtain the event camera intrinsics from the reconstructed frames. The first deep neural network to provide such utility is the E2VID network \cite{e2vid} where E2Calib \cite{eventcalib_e2vid} was proposed utilizing E2VID for the purpose of sensor calibration. Consequently, several improved networks have been developed: 1- Spade-E2VID by Cadena et al. \cite{spade_e2vid} and 2- FireNet by Scheerlinck et al. \cite{event_firenet}. However, the reconstructed images can suffer from artifacts as the sensor sweeps the calibration pattern degrading the feature extraction and calibration performance. This also shows that traditional image-based perception algorithms do not maximize the potential of event-based vision. Since traditional image processing techniques for event camera calibration do not provide the sufficient calibration accuracy, Huang et al. \cite{ecalib_iros} proposed a multi-segment based optimization to obtain the sensor intrinsics directly from events using the asymmetric circle grid for its robustness against defocused scenes \cite{circles_focus}. The approach associates events to their respective circles using density-based spatial clustering (DBSCAN) and introduces soft and hard feature extraction methods to differentiate events generated by the calibration pattern circles from background noise. The incorporated feature extraction methods rely on circle fitting for each cluster to pass events along the fitted circles to a multi-segment optimization problem to obtain the intrinsics. The drawback of such method is the presence of noisy events close to the observed circles, which affect the circle fitting accuracy leading to false detections and degraded calibration performance. In addition, the method utilizes the raw events as the control points for the calibration optimization instead of the circle centers of the circle targets. Hence, sub-pixel localization accuracy of the control points is not achieved with the edge events, which is directly correlated with degraded calibration accuracy.

\begin{figure*}[!h]
    \centering
    \includegraphics[keepaspectratio=true,scale=0.475, width=\linewidth]{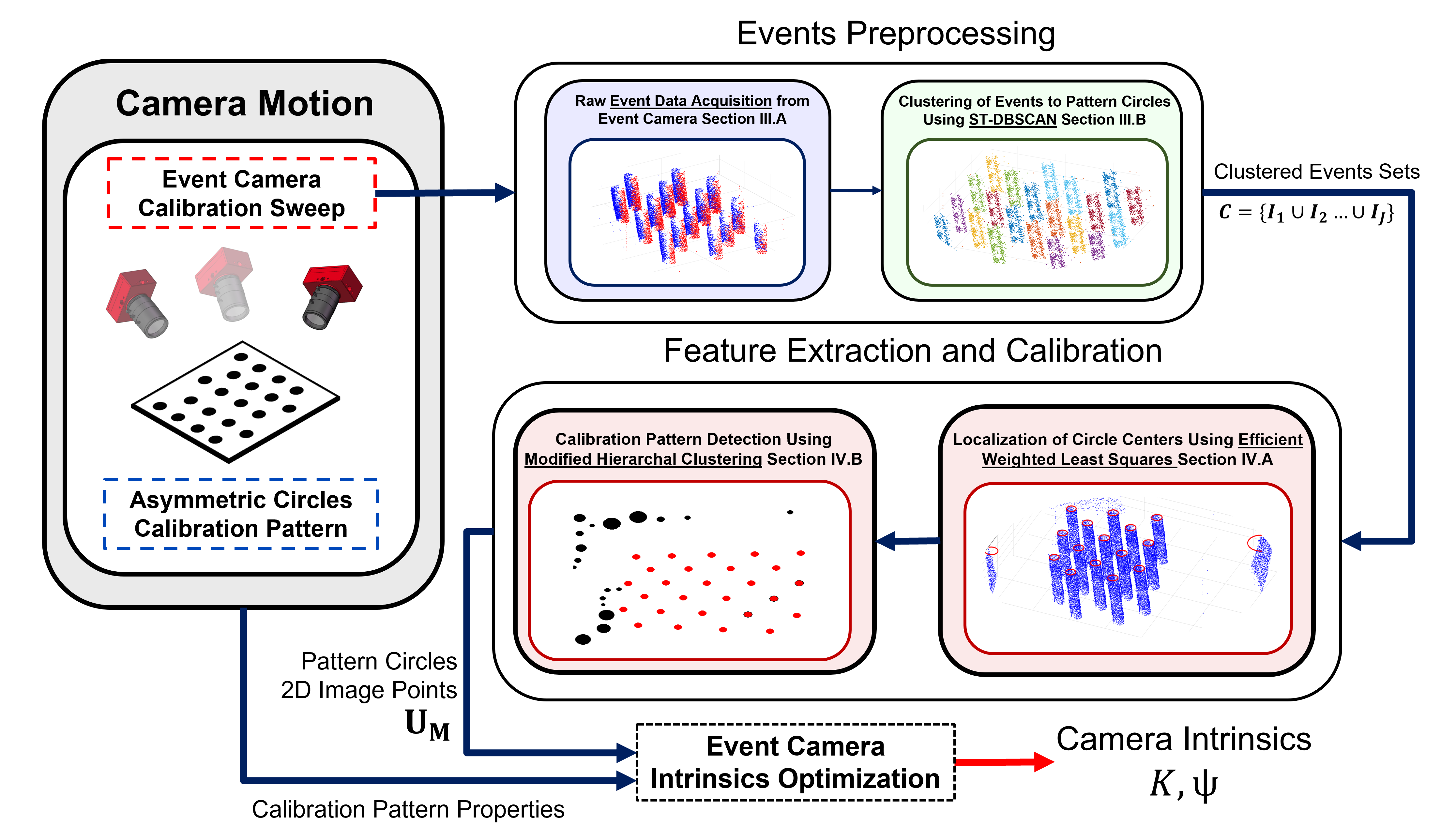}
    \caption{A high-level block diagram of the proposed calibration method. First, the raw events $\mathbf{E}_{k}$ are acquired from the event camera and are clustered to their respective circles to the clustered event sets $C = \{ {I_1, I_2, ... I_{J}} \}$ using ST-DBSCAN. Second, the centers of the circles are robustly extracted by means of eRWLS and the pattern is distinguished from the background using modified hierarchical clustering. Finally, the calibration optimization is performed using the predicted circles 2D image points $\mathbf{U}_{M}$ to obtain the event camera intrinsics.}
    \label{fig:overview}
\end{figure*}

\subsection{Structure of The Article} \label{subsection:structure}
The rest of the article is organized as follows. Section \ref{section:overview} provides a general overview on the proposed calibration framework. Section \ref{section:preprocess} outlines the events preprocessing step to assign events to their corresponding pattern circles in an unsupervised fashion. Section \ref{section:extraction} outlines the theory behind our novel feature extraction method of the calibration targets. Experimental validation of our calibration tool is provided in section \ref{section:exp} and section \ref{section:conc} presents conclusions and future aspects of the developed work.


\section{Framework Overview} \label{section:overview}
\begin{figure}[t]
\center
\includegraphics[width=0.45\textwidth]{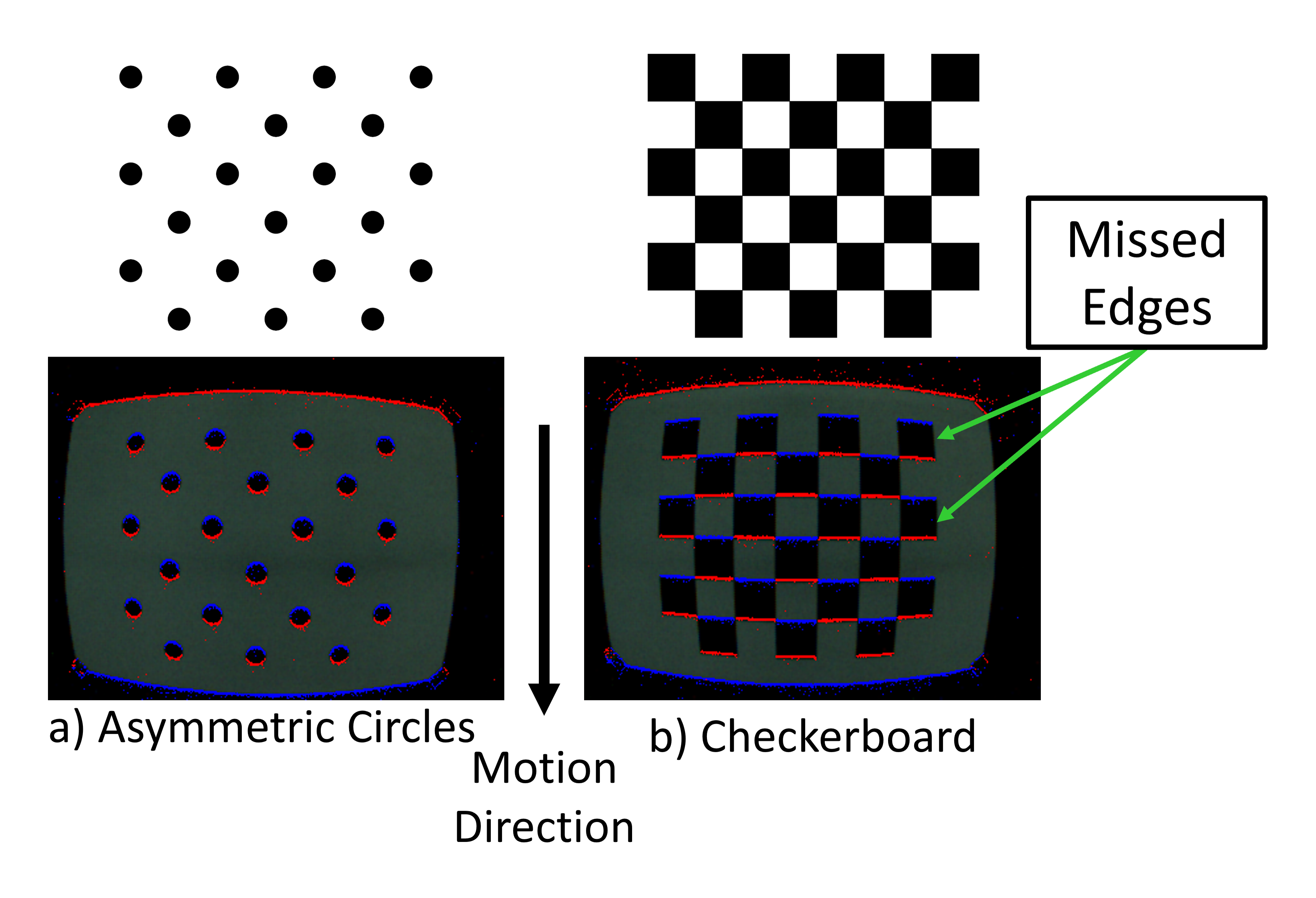}
\caption{Events generated for an event camera observing a) asymmetric circles pattern and b) checkerboard. Notice that the checkerboard edges parallel to camera motion do not fire events, while events for circular features are motion invariant.}
\label{fig:checkers_circles}
\end{figure}

An illustrative block diagram of our proposed calibration approach is shown in Fig. \ref{fig:overview}, involving a moving event camera observing an asymmetric circle grid. We chose the asymmetric circle grid as the calibration pattern for two reasons. First, the asymmetric circles calibration pattern is robust against out of focus scenes \cite{circles_focus}. Second, the features generated from the circles are motion invariant, unlike checkerboards and AprilTag Markers where edges parallel to camera motion do not fire events \cite{neuro_servoing}, see Fig. \ref{fig:checkers_circles}. As the event camera sweeps the calibration pattern of known $M$ circles, the fired events need to be associated to their corresponding circles. Thus, we devise spatiotemporal density-based clustering with applications to noise (ST-DBSCAN) \cite{dbscan} to cluster the asynchronous events to their respective circular features, outlined in section \ref{subsection:dbscan}. Unlike DBSCAN \cite{dbscan_orig} which ignores the timestamps of the events, ST-DBSCAN performs optimal event-based clustering due to its capability to operate on spatiotemporal data and adhere to the sparse output of the event camera. It is worth mentioning that ST-DBSCAN requires predefined parameters $\epsilon_{s}$ and $\epsilon_{t}$, defined as the neighborhood search radius, which require tuning for different event camera models. To circumvent this limitation, the asynchronous events are normalized with respect to the sensor resolution prior to ST-DBSCAN. This is essential to perform clustering in a unified spatiotemporal domain and ensure consistent clustering performance even for different sensor sizes.

After successfully associating the events to their respective circles, the image centers of the calibration targets need to be extracted with sub-pixel localization accuracy to optimally calibrate the sensor. Such feature is not obtained by utilizing the raw events as control points for calibration optimization. A naive approach is to rely on ST-DBSCAN cluster centers as the feature points of the calibration targets. However, due to the inherent noisy nature of the event camera, the cluster centers drift away from the actual circle centers. Even though ST-DBSCAN is characterized with robustness to noise, noisy events can still be mapped to the circle targets, especially when they are close to the real events. On the other hand, the observed circular features form dense slanted cylinders in spatiotemporal space due to monotonically increasing event stamps, preventing the applicability of image-based feature extraction methods. To extract the image points of the pattern circles, each clustered set of events is first fed to the efficient weighted least squares (eRWLS) algorithm, discussed in section \ref{subsection:eRWLS}, where cylinders are robustly fitted to the events clusters while penalizing the noisy events close to the observed circles. It is also essential to highlight that eRWLS assumes the probability density function (PDF) of the residual vector is dominated by a normal distribution. Accordingly, eRWLS efficiently reweights the residual vector while penalizing the outlier weights iteratively until convergence.

Finally, since background clutter can be observed, a modified hierarchical clustering algorithm, detailed in section \ref{subsection:hier}, is devised to extract the calibration pattern in the image plane apart from the background given the geometric properties of the asymmetric circle grid. Note that we employed the hierarchical clustering algorithm described in \cite{cut_off}. However, the algorithm is not directly applicable since hierarichal clustering on the detected circle centers fails since they do not reflect the geometric representation of the calibration pattern. Thus, we transform the input data to conform to the uniform geometric representation of the circles in the calibration pattern. This adjustment ensures clear differentiation from background clutter.

As soon as the image centers are identified, they are fed to the calibration optimizer along with the pattern geometric properties to obtain the sensor intrinsic parameters.

\section{Events Preprocessing} \label{section:preprocess}
\subsection{Event-Based Vision} \label{subsection:nvs}
The event camera is a bio-inspired technology comprising of an array of photodiode pixels triggered by log light intensity variations generating photocurrents represented by binary, asynchronous \textit{events}. Each event is spatially and temporally stamped as

\begin{equation}
    e_k\doteq(\bm{x}_k,t_k,p_k),
\end{equation}
where \(\bm{x}\in Z_{w}\times Z_{h}\) is $(x, y)_{k}$ defined as the pixel location, $Z_{w} = \{ 0,1,\cdots,w-1 \}$ and $Z_{h} = \{ 0,1,\cdots,h-1 \}$ with $W$ and $H$ represent the sensor resolution, \(t\) is the timestamp, and \(p\in\{-1,+1\}\) is the polarity of the event for a decrease and increase in the light intensity, respectively.  On the other hand, even though the event camera output is dissimilar to frame-based imaging sensors, they utilize identical optics where the standard perspective projection model is followed by event cameras defined by

\begin{equation}
        s\begin{bmatrix}
        u \\
        v \\
        1
    \end{bmatrix} = K \underbrace{\begin{bmatrix} R \text{  } T
\end{bmatrix}}_{{}^{P}_{C}\mathcal{T}} \begin{bmatrix}
x \\
y \\
z \\
1
\end{bmatrix},
\label{eq:projection}
\end{equation}

\noindent where $x$, $y$, $z$ represent the world coordinates of a point in 3D space mapped to image coordinates $u$ and $v$. ${}^{P}_{C}\mathcal{T}$ is the transformation from world, i.e. calibration pattern, coordinates to the event camera frame of reference defined by the rotation matrix $R$ and translation vector $T$. $K$ is the intrinsic matrix comprising of the focal lengths $f_{x}$ and $f_{y}$, and principal point $[u_{0}, v_{0}]$, obtained through sensor calibration. In addition to the intrinsic matrix $K$, lens distortion coefficients $\Psi$ also need to be computed when calibrating event cameras defined by

\begin{equation}
    \Psi = (k_{1}, k_{2}, p_{1}, p_{2}, k_{3})
\end{equation}

\begin{figure}[t]
\center
\includegraphics[scale=0.25]{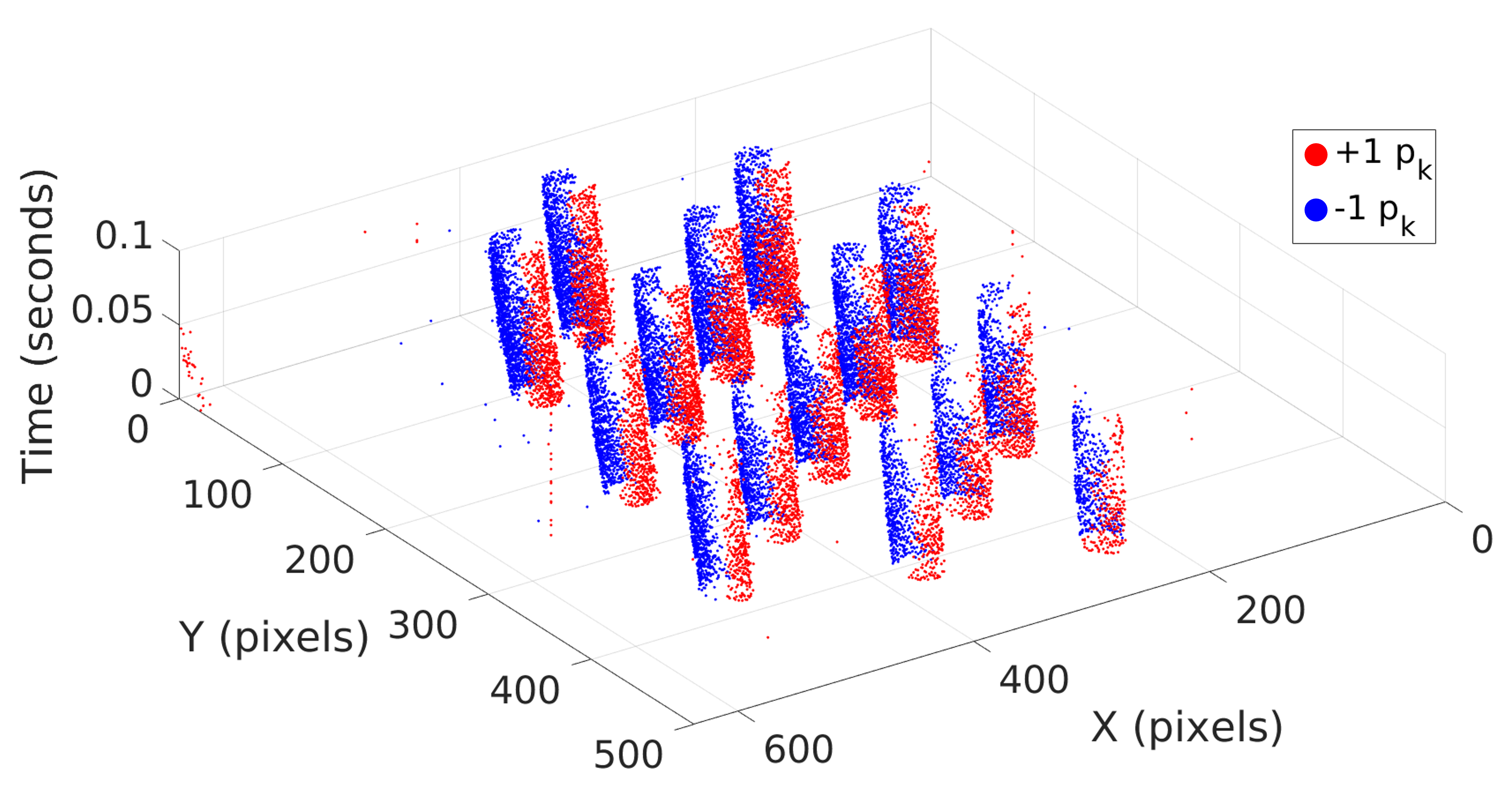}
\caption{Accumulated spatiotemporal window for an event camera observing the circles pattern. Notice that the circle targets form dense slanted cylinders due to the monotonic increase in time.}
\label{fig:cylinders}
\end{figure}

\begin{figure}[b]
\center
\includegraphics[scale=0.325]{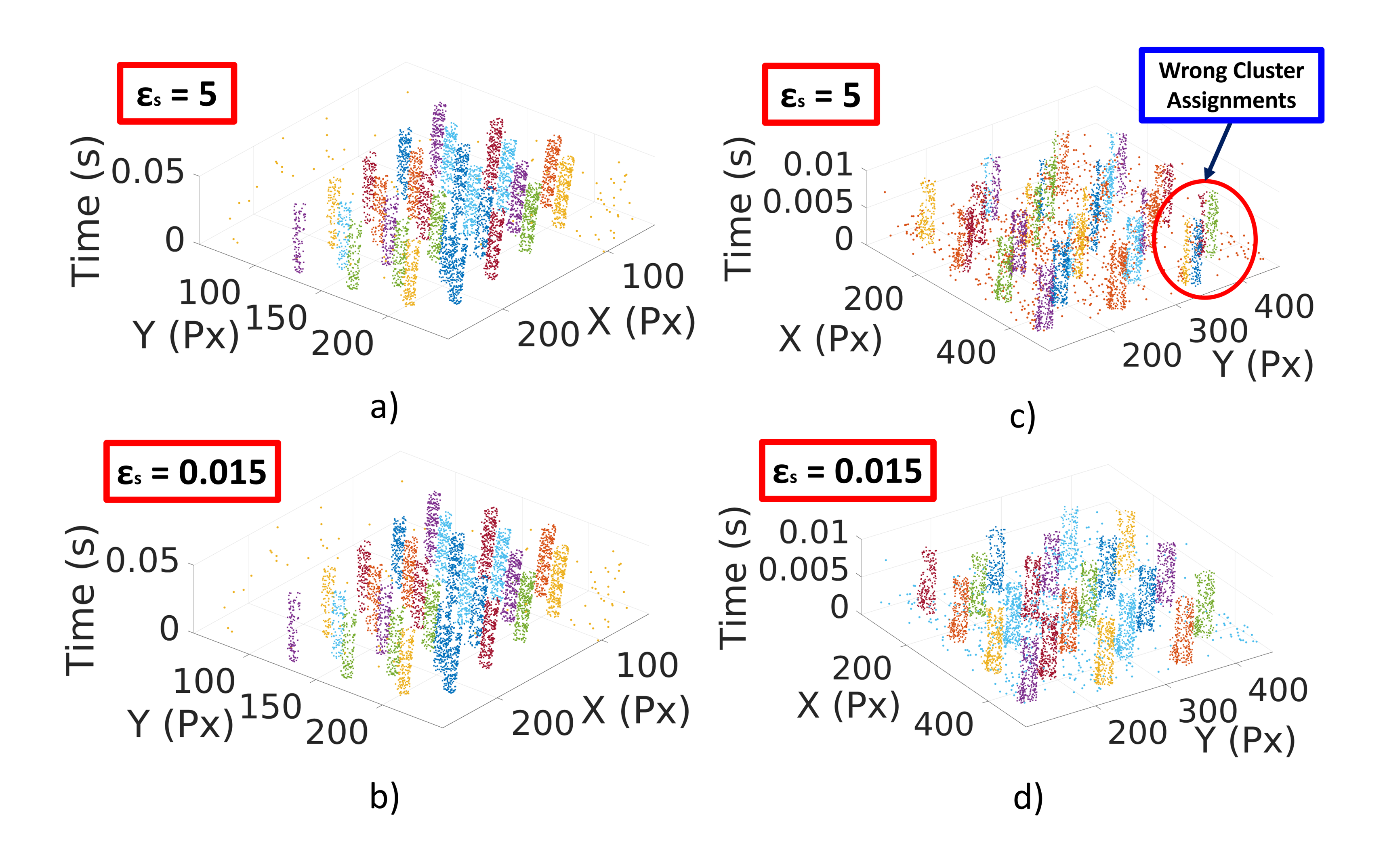}
\caption{a), b) and c), d) ST-DBSCAN on $e_{k}$ and $\hat{e}_{k}$ for DAVIS346 and DVXplorer, respectively. A fixed $\epsilon_{s}$ on $\hat{\mathbf{E}}_{k}$ maintains the desired clustering performance.}
\label{fig:normalized_plot}
\end{figure}

\noindent where $(k_{1}, k_{2}, k_{3})$ are the radial distortion coefficients, while $(p_{1}, p_{2})$ are the tangential distortion coefficients. Both $K$ and $\Psi$ are optimally acquired with accurate sensor calibration and are necessary for robust 3D understanding of the scene.

\subsection{Unsupervised Clustering of Events} \label{subsection:dbscan}

Assume that we are given a set of events $\mathbf{E}_{K}$ at time $t_{K}$ with elements $e_k$ for an event camera observing the calibration grid, where \(k\in\{1,...,K\}\) and $t_{K} > t_{1}$. The events corresponding to the observed circles form dense slanted cylinders, see Fig. \ref{fig:cylinders}. As is the case of conventional cameras, the observed circles in the calibration pattern need to be identified from one another to calibrate the sensor, where each $e_k$ needs to be preprocessed and associated to its corresponding circle in the calibration grid.

We utilize density-based spatiotemporal clustering of applications with noise (ST-DBSCAN) \cite{dbscan} algorithm to match events to their correspondent circles for to its computational efficiency and capability to adhere to the asynchronous, spatiotemporal nature of the event camera. ST-DBSCAN attempts to find dense groups of events by finding their nearest neighbors in a cylindrical region defined by the spatial radius $\epsilon_{s}$ and height $\epsilon_{t}$. Accordingly, events are iteratively assigned to their clusters if their relative euclidean distance falls below $\epsilon_{s}$ and $\epsilon_{t}$, otherwise they are regarded as noise. Obviously, ST-DBSCAN is more optimized to cluster events in spatiotemporal windows compared to DBSCAN, since a cylindrical search region is performed for clustering instead of the circular search area by DBSCAN.

Intuitively, $\epsilon_{s}$ is not unique when tackling different sensor sizes and calibration pattern variants of different geometric properties (i.e. height, width, and diagonal spacing). To circumvent this limitation, the spatial and temporal stamps of $e_{k} \in \mathbf{E}_{k}$ are normalized with respect to the sensor resolution ($H \times W$) and the spatiotemporal window time step size $\Delta_{k}$ as $\hat{e}_{k} = (\frac{x}{W} \text{ } \frac{y}{H}, \frac{t_{k} - t_{1}}{\Delta_{k}})$, where $\hat{e}_{k}$ are the normalized events. Consequently, the input to ST-DBSCAN are the normalized events. Normalizing the events' spatial and timestamps makes the clustering resolution agnostic, independent of the accumulation time step size, and eventually a fixed $\epsilon_{s}$ is utilized for different sensor models. Fig. \ref{fig:normalized_plot} shows the clustering performance on $\mathbf{E}_{K}$ and $\hat{\mathbf{E}}_{K}$ both comprising of 4000 events for two sensor models of different resolutions, DAVIS346 and DVXplorer. Note that the same value of $\epsilon_{s}$ is identical for $\hat{\mathbf{E}}_{k}$ as the events are unified under the same domain range for both sensors while ST-DBSCAN fails on the DVXplorer raw events.

The output of ST-DBSCAN are the sets of clustered events $C = \{ {I_1, I_2, ... I_{J}} \}$, with $\mathbf{\hat{E}}_K$ partitioned as $\mathbf{\hat{E}}_K = I_1 \cup I_2 \cup ... \cup I_J$, where $I_j \in C$ and $J$ are the total number of clusters. Note that the spatial coordinates are in the original image plane and not the normalized coordinates. This is essential to extract the circle centers in the image plane for calibration and not in normalized coordinates. In addition, the total number of events utilized to formulate a spatiotemporal window of clustering is a hyperparameter. We avoided forming spatiotemporal windows based on fixed time intervals since at certain instances a very small relative motion is encountered and no sufficient features can be extracted from $\hat{e}_{k}$. In addition, large number of events can lead to curved cylinders generated from the observed circles. This tends to have insignificant effect on the calibration accuracy due to the weighing factor placed in the eRWLS cost function. Yet, the number of events is considered a hyperparameter to ensure the highest calibration accuracy can be obtained. In our experiments, we utilized at least 4000 events to cluster $\mathbf{\hat{E}}_{K}$ with a time step of 33 ms between the spatiotemporal windows during experiments. This was sufficient to obtain the required calibration accuracy. On the other hand, $\epsilon_{s}$ and $\epsilon_{t}$ are set to 0.015 providing optimal clustering performance. It is also important to highlight that the event polarities are ignored in the clustering algorithm.

\begin{figure}[b]
\center
\includegraphics[scale=0.265]{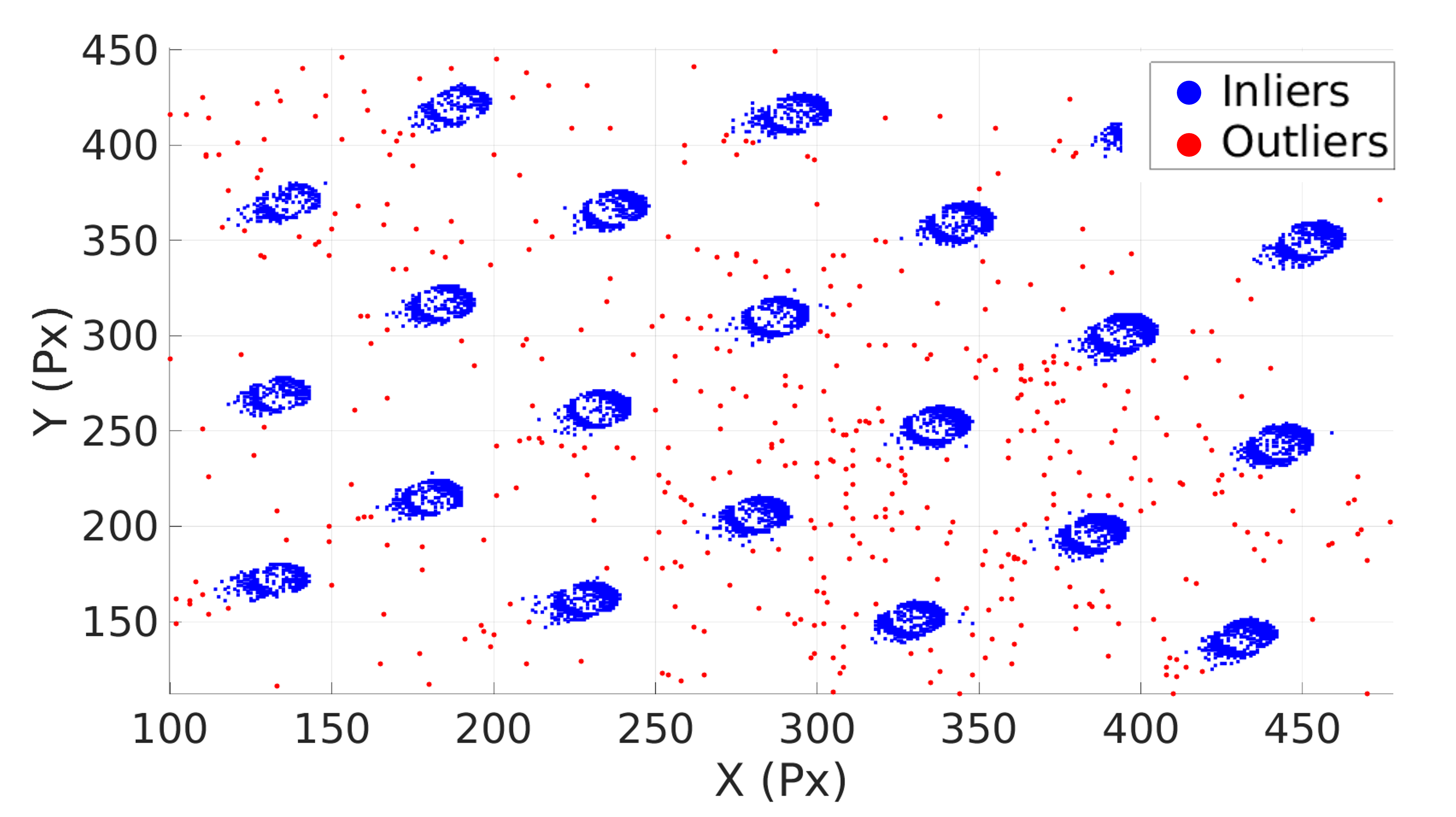}
\caption{The clustered events from ST-DBSCAN. Noisy events (outliers) are still evident close to the observed circle targets. Note that the figure is for illustrative purposes and events extend in the time dimension.}
\label{fig:dbscan_outliers}
\end{figure}

\section{Feature Extraction and Calibration} \label{section:extraction}

\begin{figure}[t]
\center
\includegraphics[scale=0.25]{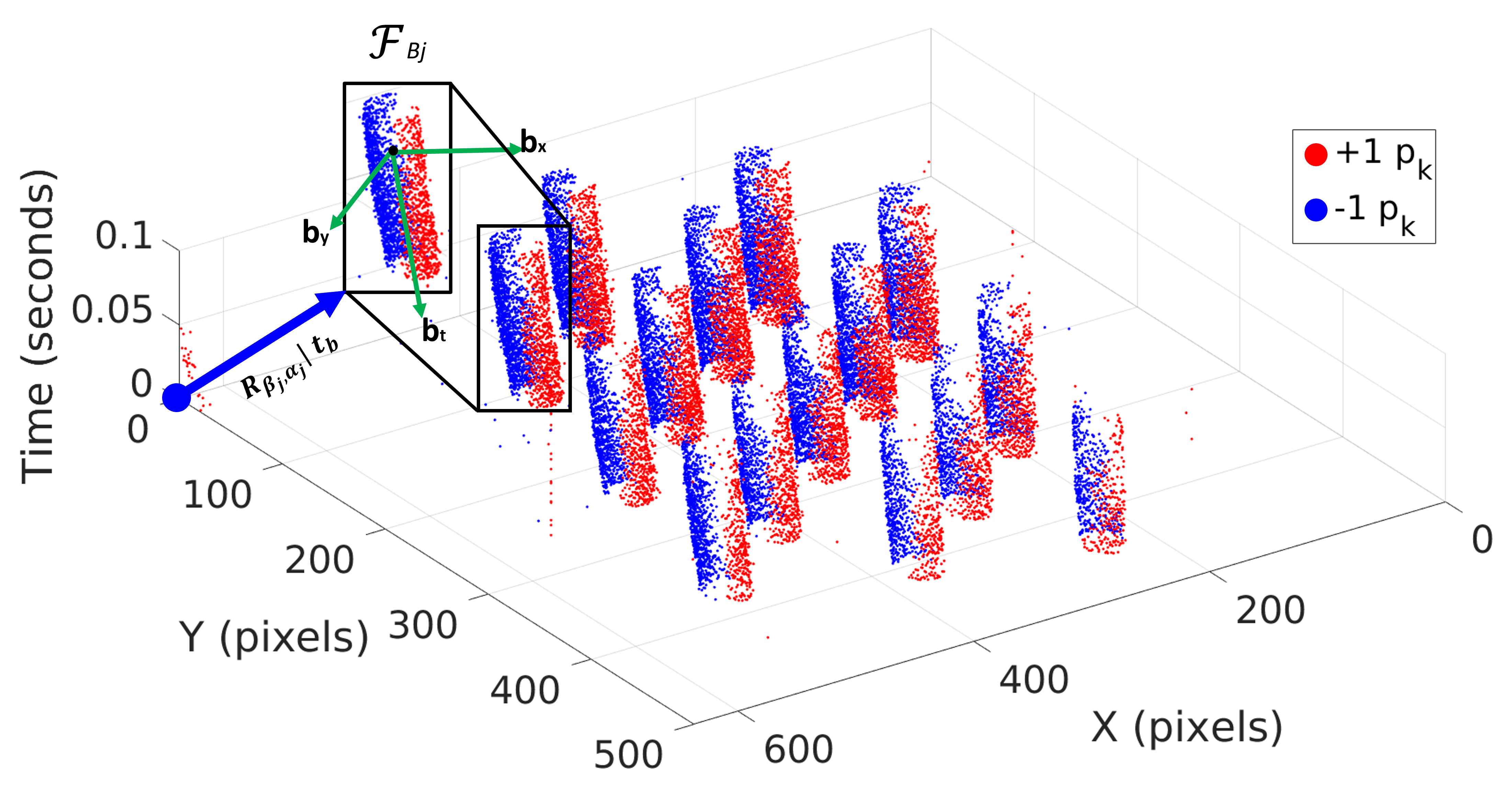}
\caption{The body frame $\mathcal{F}_{bj}$ defined for each cluster in the spatiotemporal window with the cylinder center defined by $\mathbf{t}_{b}$.}
\label{fig:sync}
\end{figure}

\subsection{Efficient Reweighted Least Squares} \label{subsection:eRWLS}
\begin{algorithm}[t]
\caption{Efficient Reweighted Least Squares}
\begin{algorithmic}[1]
\Inputs{Clustered events sets $C = \{ I_{1}, I_{2}, ..., \, \ I_{J} \}$} \\
\textbf{Outputs}: {Cylinder parameters: $\mathbf{\Omega}_{j}$}
\Initialize{\strut $\mathbf{\Omega}_{j} = \omega_{j} = [\bar{r}_{j}, \bar{u}_{j}, \bar{v}_{j}, \beta_{j}=0, \alpha_{j}=0]$}
\For{$j$ = $1$ to $J$}
\While{Not converged}
\State \textbf{Find} $R_{\beta,\alpha}$ and $\mathbf{t}_{b}$
\State \textbf{Transform} $e_{k}$ using Eq. \ref{eq:transformation}
\State \textbf{Find} $\boldsymbol{\xi}$ and $\mathbf{w}_{j}$, using Eqs. \ref{eq:residuals} and \ref{eq:weights}
\State \textbf{Find} $\sum_{k} w_{k}\boldsymbol{\xi}^{2}$
\State \textbf{Evaluate} $\mathbf{J}$ and \textbf{Step} $\mathbf{\Omega_{j}}$
\EndWhile
\EndFor
\end{algorithmic}
\label{alg:circle_detect}
\end{algorithm}

After obtaining the sets of clustered events $C$, each of the clustered sets $\{ I_1, I_2, ... , I_{J} \}$ are fed to eRWLS to efficiently fit and extract the center of the dense cylinders formed by the observed circles grid. We first highlight the motivation of eRWLS by showing the remaining presence of outliers, where the cluster centers drift away from the true cylinder center, see Fig. \ref{fig:dbscan_outliers}. Similarly, utilizing ordinary least squares (OLS) tends to be affected by the presence of noise leading to degraded detection performance of the circle centers. In contrast, eRWLS attempts to predict the desired centers even in the presence of persisting sensor noise. Algorithm \ref{alg:circle_detect} defines the algorithmic procedure of eRWLS for estimating the cylinder centers. It is also worth mentioning that the predicted centers by eRWLS are not the actual image points to be considered for the calibration optimization. Instead, the estimated centers are fed to the modified hierarchical clustering algorithm for detecting the circles grid before the calibration optimization.

\begin{figure}[b]
\center
\includegraphics[scale=0.45]{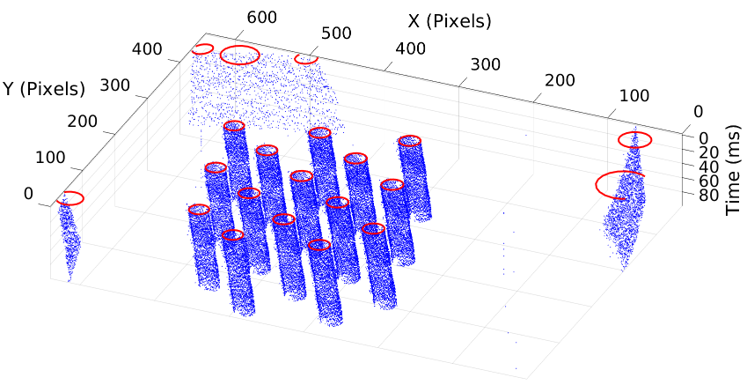}
\caption{Extracted cylinder centers by eRWLS. Notice that all centers are synchronized at $t_{ref} = t_{1}$.}
\label{fig:cap}
\end{figure}

As depicted in Fig. \ref{fig:cylinders}, the events for the calibration targets follow a cylinder rotated around the $x$ the $y$ axes of the spatiotemporal window $\mathcal{F}_{S}$ with angles $\beta_{j}$ and $\alpha_{j}$, respectively. We define a body frame for each of cylinder \(\mathcal{F}_{Bj}\) of basis \( \bm{[b_x, b_y, b_t]} \), where \(\bm{b_t}\) resembles a monotonically increasing time dimension, see Fig. \ref{fig:sync}. Accordingly, we define the transformation, $^{B}_{S}{\mathcal{T}}_{j}$ described by the rotation $R_{\beta_{j},\alpha_{j}}$ and $\mathbf{t}_{b} = [u_{j}, v_{j}, t_{ref}]$ transforming the events coordinates from $\mathcal{F}_{S}$ to \(\mathcal{F}_{Bj}\) as

\begin{equation}
    {}^{j}\Tilde{e}_{k} = \begin{bmatrix}
    \Tilde{x}_{k} \\
    \Tilde{y}_{k} \\
    \Tilde{t}_{k}
    \end{bmatrix}
    = \underbrace{
    \begin{bmatrix}
    R_{\beta_{j},\alpha_{j}} & \mathbf{t}_{b} \\
    0 & 1
    \end{bmatrix}}_{^{B}_{S}{\mathcal{T}}_{j}}
    \begin{bmatrix}
    x_{k} \\
    y_{k} \\
    t_{k} \\
    1
    \end{bmatrix},
    \label{eq:transformation}
\end{equation}

\noindent where ${}^{j}\Tilde{e}_{k}$ are the events defined in \(\mathcal{F}_{Bj}\), $R_{\beta_{j},\alpha_{j}}$ is the cylinder's orientation, and $\mathbf{t}_{b}$ is the translation vector pointing from $\mathcal{F}_{S}$ origin to the cylinder center. Note that the cylinder height is constrained at reference time $t_{ref} = t_{1}$ and is not optimized. Meaning, all cylinder centers are set to their caps at $t_{1}$, see Fig. \ref{fig:cap}. This is of paramount importance to ensure that all cylinder centers are synchronized in time. Note that any reference time can be used ranging from $t_{1}$ to $t_{k}$ as long as all the fitted circles are synchronized to the chosen $t_{ref}$.

To optimize the other the cylinder parameters, $\boldsymbol{\Omega_{j}} = [r_{j}, \text{ } u_{j}, \text{ } v_{j}, \text{ } \beta_{j}, \text{ } \alpha_{j}]$, a nonlinear least square problem can be constructed to minimize the squared residuals $\boldsymbol{\xi}$ formulated as

\begin{figure}[h]
\center
\includegraphics[scale=0.33]{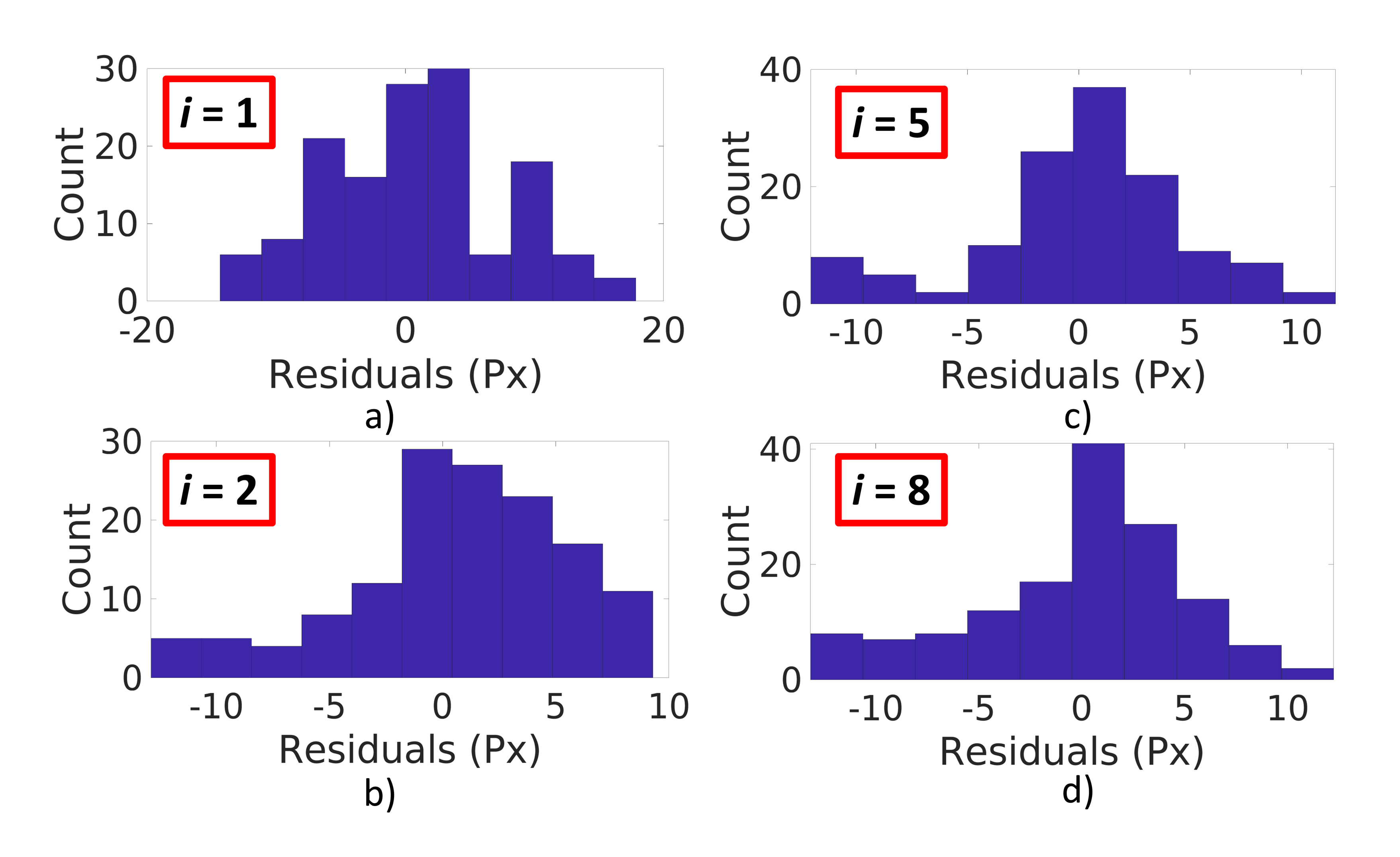}
\caption{The PDF for $\xi$ at consecutive iterations. The PDF starts to slowly converge to a normal distribution progressing from a) iteration 1 to d) iteration 8.}
\label{fig:pdf}
\end{figure}

\begin{equation}
    \boldsymbol{\Omega_{j}}^{*} = \argmin_{\boldsymbol{\Omega}_{j}}\sum_{k=1}^{K}\biggl( \underbrace{(\Tilde{x}_{k} - u_{j})^{2} + (\Tilde{y}_{k} - v_{j})^{2} - r_{j}^{2}}_{\boldsymbol{\xi}}\Biggl)^{2}.
    \label{eq:residuals}
\end{equation}

\noindent where $u_{j}$ and $v_{j}$ are the pixel coordinates of the cylinder's center, while $r_{j}$ is the cylinder radius. However, the noise in event stream as shown in Fig. \ref{fig:dbscan_outliers}, will introduce outliers to the optimization framework which can significantly affect the optimization results. A method that provides solutions to the least squares problem that is robust against noise is the weighted least squares, but the weights for each element in the residual vector need to be provided in prior. On the other hand, iteratively reweighted least squares (IRLS) \cite{irls} can iteratively recompute the weight vector but suffers from convergence difficulties and leads to undesired computational complexities. To alleviate these restrictions, we show that the probability density function (PDF) of $\boldsymbol{\xi}$ is dominated with a Gaussian distribution and the weights $\mathbf{w}$ can be implicitly updated by a normally distributed vector as

\begin{equation}
    \mathbf{w}_{j} = \dfrac{1}{\sigma \sqrt{2 \pi}} \exp{(-\frac{\boldsymbol{\xi} \mu^{2}}{{2\sigma^{2}}})},
    \label{eq:weights}
\end{equation}

\noindent where $\mu$ and $\sigma$ are the mean and standard deviation of $\boldsymbol{\xi}$. Accordingly, the weights for the outliers are severely weakened and the cost function in Eq. \ref{eq:residuals} is modified to

\begin{equation}
    \mathbf{\Omega}^{*}_{j} = \argmin_{\mathbf{\Omega}_{j}}\sum_{k} w_{k}\boldsymbol{\xi}^{2},
    \label{eq:cost}
\end{equation}

\noindent where $w_{k}$ are the elements of $\mathbf{w}_{j}$ and Eq. \ref{eq:cost} is iteratively minimized using the Levenberg–Marquardt algorithm \cite{lm} to compute $\Omega_{j}$. It is important to highlight that by implicitly computing $\mathbf{w}$ of the residuals, the optimization process not only optimizes $\Omega_{j}$ but inherently maximizes the likelihood of the PDF, as demonstrated in Fig. \ref{fig:pdf}. Notice that at the first iteration, the PDF is random but as the algorithm converges (i.e. iterations 5 and 8), the PDF is dominated by a normal distribution centered around $\mu = 0$.

One important aspect of the optimization process is the initial condition $\omega_{j}$ of $\Omega_{j}$. To maintain stability and fast convergence, we define $\omega_{j}$ to be the median of the transformed spatial stamps ${}^{j}\Tilde{e}_{k}$ along with their norm $\bar{r}_{j}$ as $\bar{u}_{j}$ and $\bar{v}_{j}$, while $\beta_{j}$ and $\alpha_{j}$ are initialized as zeros. The optimization is run with a variable differential step computed from the jacobian $\mathbf{J}$ \cite{lm}, which has been found analytically using MATLAB symbolic toolbox \cite{matlabsymbolic}.

\subsection{Modified Hierarichal Clustering} \label{subsection:hier}

\begin{figure}[b]
\center
\includegraphics[scale=0.275]{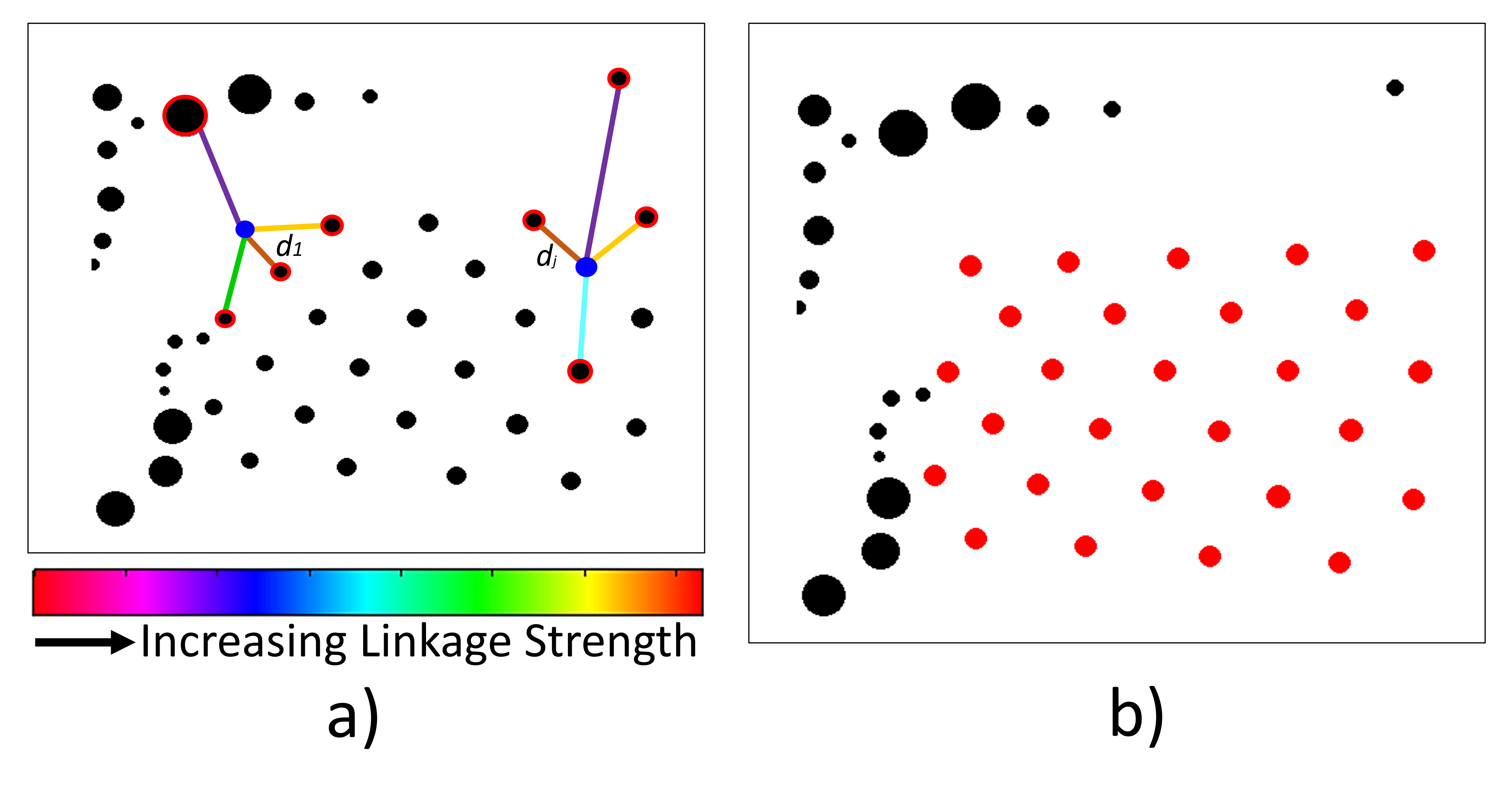}
\caption{a) All detected image points by the event-based circle detector with linkages formed between the clusters. b) Our modified hierarchical clustering method extracts the actual image points (red) apart from the false detections (black).}
\label{fig:points}
\end{figure}

The synchronized cylinder centers $\mathbf{U} = \{(u_{j}, v_{j})\}_{j=1}^{J}$ include the actual centers of the pattern circles in addition to false detections of the background clutter, see Fig. \ref{fig:cap}. This occurs when the sensor observes the scene background as it sweeps the calibration pattern during the data collection step. Hence, a subset of $\mathbf{U}$,  $\mathbf{U}_{M} \subset \mathbf{U}$ of $M$ elements corresponding to the total number of pattern circles, needs to be extracted to be fed the calibration optimization. We devise a modified hierarchical clustering \cite{cut_off} algorithm, outlined in algorithm \ref{alg:mhc}, that robustly identifies $\mathbf{U}_{M}$ from $\mathbf{U}$ as the desired image points for calibration optimization and the rest of the $J-M$ centers as outliers, where $J > M$. It is worth mentioning that the clustering algorithm in \cite{cut_off} is utilized and we rather modify the input such that it is represented by the uniform geometrical representation of the circles in the calibration pattern to distinguish it from the background clutter.

We create a single linkage per each synchronized center to the centers with the nearest relative euclidean distance $d_{j}$ with its index as

\begin{equation}
    j\prime^{*} = \argmin_{j\prime=1,2,...J\not= j}\Vert (u_{j}, v_{j})- (u_{j\prime}, v_{j\prime}) \Vert ^{2}, \\
    \label{eq:jprime}
\end{equation}

\noindent and the obtained $j\prime$ is utilized to find the corresponding euclidean distance linkage as

\begin{equation}
    d_j = \Vert (u_j, v_j) - (u_{j'}, v_{j'})\Vert ^{2}.
    \label{eq:dj}
\end{equation}

\noindent This forms the set $\mathbf{D} = \{ d_{1}, d_{2}, \text{ } \cdots, d_{J} \}$ and the corresponding cylinder radii set as $\mathbf{r} = \{ r_{1}, r_{2},...,r_{J} \}$. Due to the uniform diagonal spacings across the circles and their identical radii, elements of $\mathbf{D}$ and $\mathbf{r}$ corresponding to the pattern circles are very close together compared to the background which are random and sparse, see Fig. \ref{fig:points}. In addition, typical camera calibration procedures involve known number of pattern circles, where the maximum cluster cut-off method \cite{cut_off} is utilized to match the detected points to the calibration grid. The cut-off threshold is equal to $M$ and $\mathbf{U}_{M}$ corresponds to the $M$ elements with the minimum variance of $\mathbf{D}$ and $\mathbf{r}$. Therefore, the variance of all unique combinations of sets $\mathbf{D}$ and $\mathbf{r}$ needs to be evaluated such that the indices associated with the minimum variance correspond to $\mathbf{U}_{M}$. We define an index set $\Phi \subset \{ 1,...,J \}$ such that $ |\Phi|=M$ used as follows

\begin{equation}
    \mathbf{S}_{\Phi} = \{ s_{j} \in \mathbf{S} \, \vert \, j\in\Phi \}. \\
    \label{eq:phi}
\end{equation}

\begin{algorithm}[t]
\caption{Modified Hierarchical Clustering}
\begin{algorithmic}[1]
\Inputs{$\mathbf{U}_{j} = \{(u_{j}, v_{j})\}_{j=1}^{J}$, $\mathbf{r} = \{r_{1},...,r_{j} \}$ } \\
\textbf{Outputs}: {Pattern Circle Centers: $\mathbf{U}_{M}$}
\State \textbf{Find} linkages $j\prime^{*}$, $d_{j}$ using Eq. \ref{eq:jprime} and \ref{eq:dj}
\State Construct $\mathbf{D} = \{ d_{1},...,d_{j} \}$
\For{$i$ = 1 to ${}_{J}C_{M}$}
\State \textbf{Extract} $\mathbf{D}_{\Phi}$, $\mathbf{r}_{\Phi}$ with $M$ elements using Eq. \ref{eq:phi}
\State \textbf{Find} $\varphi_{i}$ using Eq. \ref{eq:mhc}
\If{$\varphi_{i} < \varphi_{-1}$}
\State $\varphi_{-1} = \varphi_{i}$
\EndIf
\EndFor
\end{algorithmic}
\label{alg:mhc}
\end{algorithm}

\noindent This is also understood as extracting all unique combinations of $M$ elements from $\mathbf{S}$, which will be utilized to extract subsets from $\mathbf{D}$ and $\mathbf{r}$. Consequently, $\mathbf{U}_{M}$ is found as

\begin{equation}
   \mathbf{U}^{*}_{i} = \argmin_{\Phi} \Vert \underbrace{\sigma\bigl( \mathbf{D}_{\Phi} \bigl) + \sigma\bigl( \mathbf{r}_{\Phi} \bigl) }_{\varphi} \Vert
   \label{eq:mhc}
\end{equation}

\noindent where $\sigma$ is the standard deviation of the set. Note that Eq. \ref{eq:mhc} is evaluated ${}_JC_M$ times and $\mathbf{U}_{M}$ is identified with index set $\Phi$ that corresponds to the minimum joint variance norm $\varphi$ of $\mathbf{D}$ and $\mathbf{r}$.

Once $\mathbf{U}_{M}$ is identified for sufficient number of pattern detections $N$, it is utilized as the control points for the calibration optimization, where the estimated parameters are the intrinsic matrix $K$ and the distortion coefficients $\Psi$. Given their initial guess, $\mathbf{U}_{M}$ is shifted by the radial distortion coefficients as

\begin{align}
    x_{i} &=  x_{r}(1+k_{1}r^{2}+k_{2}r^{4}+k_{3}r^{6}) \\
    y_{i} &=  y_{r}(1+k_{1}r^{2}+k_{2}r^{4}+k_{3}r^{6}).
\end{align}

\noindent where \( (x_i, y_i) = (u_i - u_0, v_i - v_0) \) and \( r= \sqrt{x_i^2+ y_i^2} \), with \( [u_0, v_0] \) being the principal point. Consequently, the tangential distortion is accounted for by

\begin{align}
    x_{r} &= x_{r,t}+[2p_{1}x_{r,t}y_{r,t}+p_{2}(r^{2}+2x_{r,t}^{2})] \\
    y_{r} &= y_{r,t}+[p_{1}(r^{2}+2y_{r,t}^{2})+2p_{2}x_{r,t}y_{r,t}].
\end{align}

\noindent The undistorted points, $u_{r,t} = x_{r,t}+u_0, v_{r,t} = y_{r,t} + v_0$ form the undistorted image points vector $\mathbf{U}_{r,t}$ and the camera parameters are obtained by minimizing the reprojection error as

\begin{equation}
    K^{*},\Psi^{*} = \argmin_{K,\Psi} \sum_{n=1}^{N} \sum_{i=1}^{M} \Vert\mathbf{U}_{r,t} - \pi(K, {}^{P}\mathbf{U},{}^{P}_{C}\mathcal{T})\Vert^{2},
    \label{eq:calib}
\end{equation}

\noindent where $N$ is the total number of pattern detections, $M$ is the total number of circles in the calibration pattern, ${}^{P}\mathbf{U}$ are the coordinates of the calibration targets defined in the pattern reference frame, and $\pi$ is the projection of ${}^{P}\mathbf{U}$ into the event camera frame, which is obtained by the camera projection matrix defined in Eq. \ref{eq:projection}. Eq. \ref{eq:calib} is minimized by the levenberg-Marquardt algorithm to obtain $K$ and $\Psi$ \cite{flexible_calib}.

\section{Experiments} \label{section:exp}

\subsection{Experimental Setup} \label{subsection:experimental_setup}

\begin{figure}[t]
\center
\includegraphics[scale=0.3]{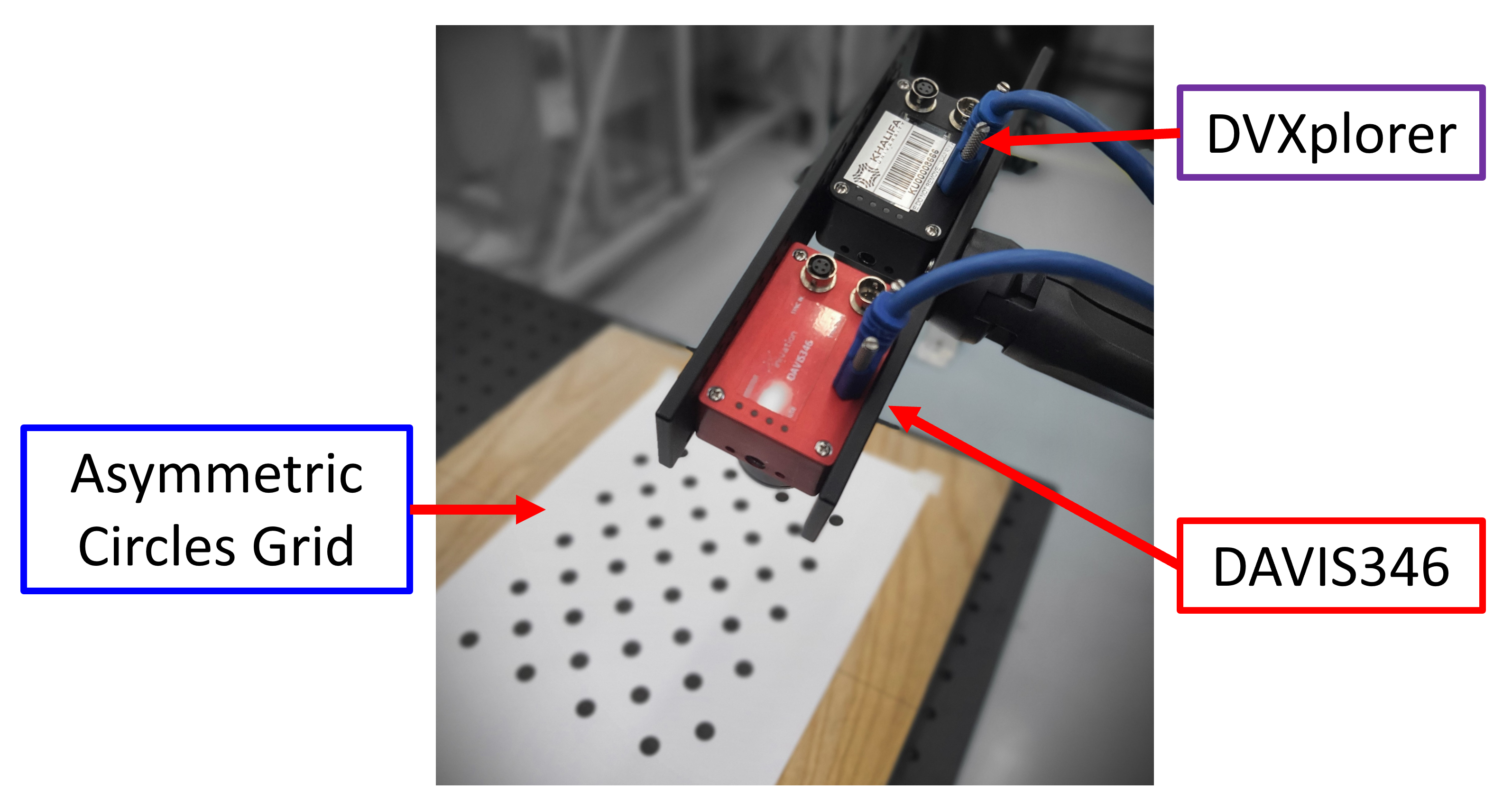}
\caption{Experimental setup for validating the proposed calibration algorithm for event camera variants observing an asymmetric circles calibration pattern.}
\label{fig:davis_dvx}
\end{figure}

We have tested the proposed algorithm in various challenging scenarios demonstrated in our \textit{ECam\_ACircles} dataset. To show that our proposed approach is resolution agnostic, event camera models with different resolutions were calibrated by the devised algorithm; the DAVIS346 $(346\times260)$ and DVXplorer ($640\times480$), see Fig. \ref{fig:davis_dvx}. Both cameras are interfaced with robot operating system (ROS) using a USB 3.0 terminal on the host computer. We have also tested our method on three calibration pattern variants of \textit{ECam\_ACircles}, of dimensions $3\times7$ and 34 mm diagonal spacing, $3\times9$ and 27 mm diagonal spacing, and $4\times11$ and 24 mm diagonal spacing. This is of paramount importance to prove that the developed calibration framework successfully calibrates event cameras regardless of the pattern geometric properties. In addition, the work of the paper was assessed in good (93.21 Lux) and poorly lighted environments (8.72 Lux) to demonstrate robustness against illumination variation and increasing noise with degraded lighting conditions. Finally, E-Calib was validated on three different paths, i.e. cone, square, and spiral, at varying speeds to test the method against harsh motion trajectories including out-of-focus scenes. Videos of the experiments are available through the following link: \url{https://youtu.be/4giQn6rt-48}.

We evaluated our method on the aforementioned experiments based on three metrics: 1) Detection success rate, 2) calibration reprojection error, and 3) positioning error of the estimated extrinsics. While comparing the calibration reprojection error and the obtained intrinsics to the DAVIS346 frames can be sufficient for validation, inaccurate intrinsics can be obtained as the solver can converge to a local minimum, even though a low reprojection error is achieved. This is usually witnessed with the lens distortion coefficients. Thus, we additionally compare the calibrated extrinsics to the ground truth pose of the camera, which is provided by OptiTrack Prime 13 motion capture system to further evaluate our method fairly. It is worth mentioning that the positioning measurements from OptiTrack and the estimated extrinsics are synchronized using soft time synchronization \cite{msg_filters}. In addition, the calibration patterns coordinates were registered to the OptiTrack frame by extrinsic calibration using the method in \cite{halwani, ayyad_drilling}.

\begin{figure}[t]
\center
\includegraphics[scale=0.35]{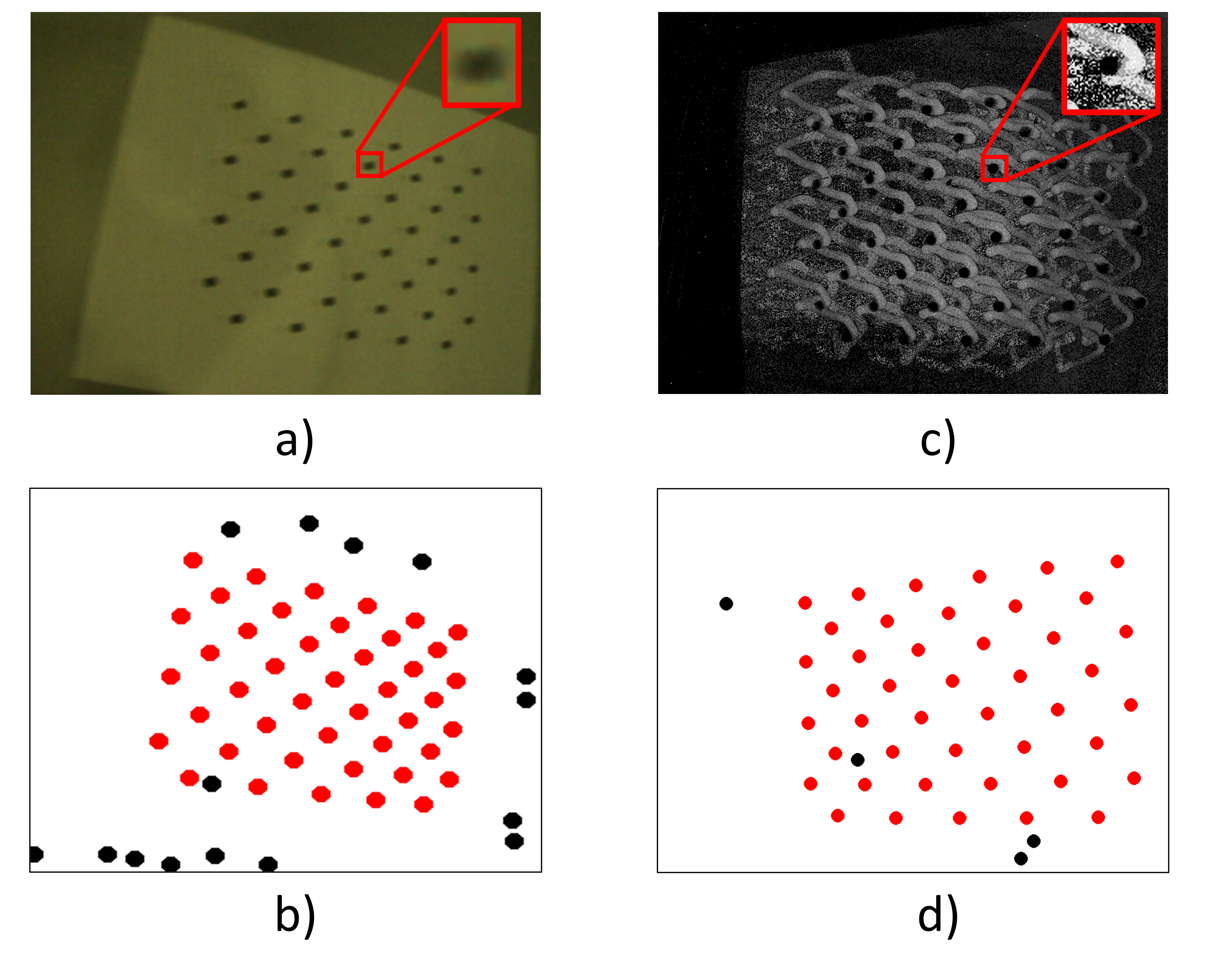}
\caption{Asymmetric circle grid detected using \cite{find_circles} on a) DAVIS and c) DVXplorer MR frame, while b) and d) represent our approach detecting a $4\times 11$ grid.}
\label{fig:detection}
\end{figure}

\begin{figure}[b]
\center
\includegraphics[scale=0.275]{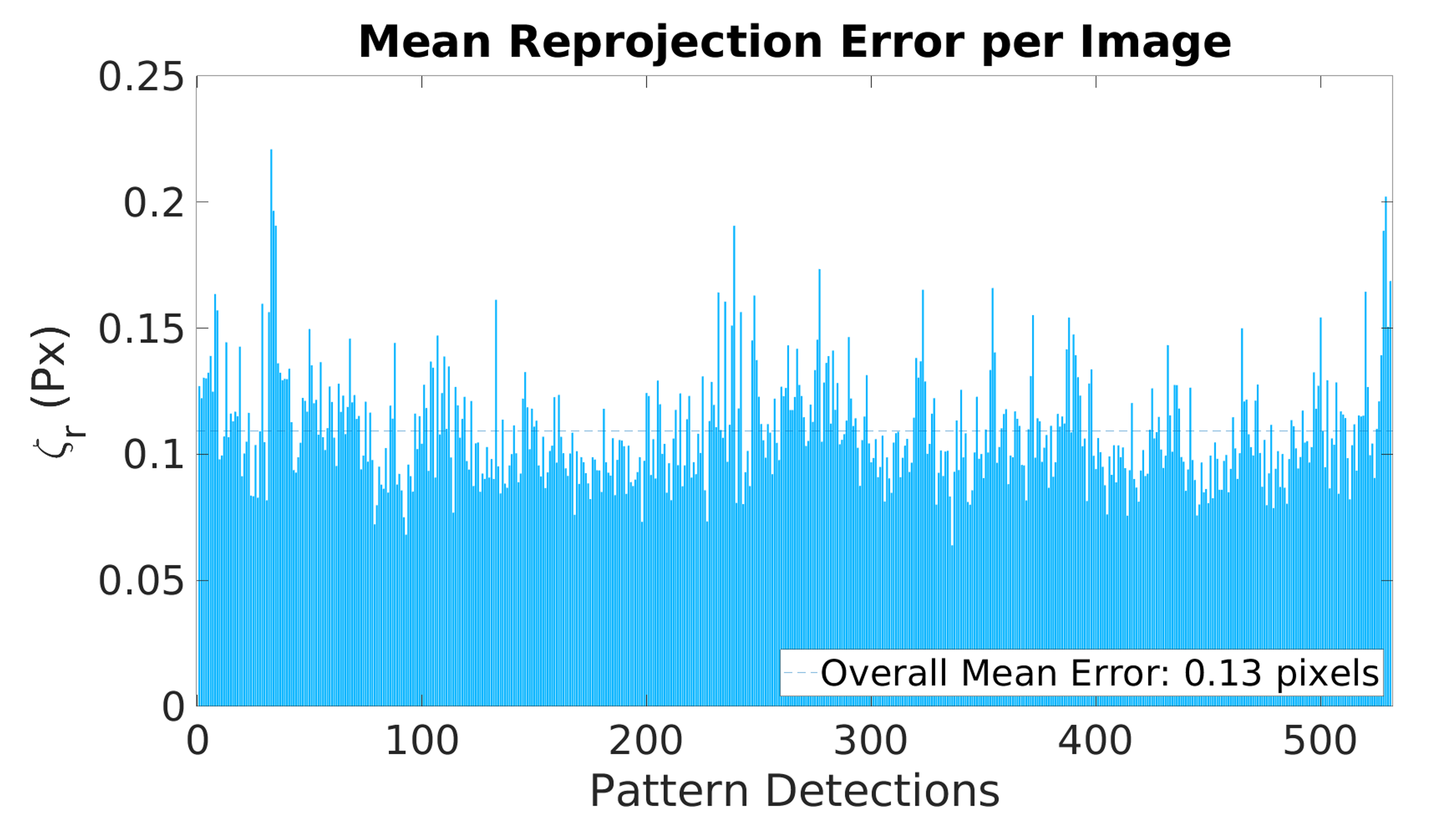}
\caption{Reprojection errors on \textit{davis\_3x9\_cone\_gdlight} dataset where $\zeta_{r} = 0.13$ pixels was obtained.}
\label{fig:error}
\end{figure}

\begin{table*}[ht]
\centering
\caption{Detection success rate and reprojection error on the cone trajectory sequences of the \textit{ECam\_ACircles} dataset.} Notice that high detection success rates and low reprojection errors are maintained even at challenging lighting conditions demonstrating robustness against noise.
\resizebox{\textwidth}{!}{%
\begin{tabular}{|c|ccccccccc|}
\hline
\multirow{3}{*}{\textbf{Metric}} & \multicolumn{9}{c|}{\textit{\textbf{Dataset}}} \\ \cline{2-10} 
 & \multicolumn{1}{c|}{\textit{\textbf{\begin{tabular}[c]{@{}c@{}}davis\_3x7\\ gdlight\end{tabular}}}} & \multicolumn{1}{c|}{\textit{\textbf{\begin{tabular}[c]{@{}c@{}}davis\_3x7\\ lowlight\end{tabular}}}} & \multicolumn{1}{c|}{\textit{\textbf{\begin{tabular}[c]{@{}c@{}}davis\_3x9\\ gdlight\end{tabular}}}} & \multicolumn{1}{c|}{\textit{\textbf{\begin{tabular}[c]{@{}c@{}}davis\_3x9\\ lowlight\end{tabular}}}} & \multicolumn{1}{c|}{\textit{\textbf{\begin{tabular}[c]{@{}c@{}}davis\_4x11\\ gdlight\end{tabular}}}} & \multicolumn{1}{c|}{\textit{\textbf{\begin{tabular}[c]{@{}c@{}}davis\_4x11\\ lowlight\end{tabular}}}} & \multicolumn{1}{c|}{\textit{\textbf{\begin{tabular}[c]{@{}c@{}}dvx\_3x7\\ gdlight\end{tabular}}}} & \multicolumn{1}{c|}{\textit{\textbf{\begin{tabular}[c]{@{}c@{}}dvx\_3x9\\ gdlight\end{tabular}}}} & \textit{\textbf{\begin{tabular}[c]{@{}c@{}}dvx\_4x11\\ gdlight\end{tabular}}} \\ \hline
\textbf{\begin{tabular}[c]{@{}c@{}}Data Sequence\\ Time (s)\end{tabular}} & \multicolumn{1}{c|}{17.04} & \multicolumn{1}{c|}{15.37} & \multicolumn{1}{c|}{27.02} & \multicolumn{1}{c|}{24.04} & \multicolumn{1}{c|}{19.07} & \multicolumn{1}{c|}{18.53} & \multicolumn{1}{c|}{15.53} & \multicolumn{1}{c|}{27.04} & 14.99 \\ \hline
\textbf{\begin{tabular}[c]{@{}c@{}}Total\\ Detections\end{tabular}} & \multicolumn{1}{c|}{438} & \multicolumn{1}{c|}{362} & \multicolumn{1}{c|}{709} & \multicolumn{1}{c|}{648} & \multicolumn{1}{c|}{437} & \multicolumn{1}{c|}{392} & \multicolumn{1}{c|}{330} & \multicolumn{1}{c|}{672} & 312 \\ \hline
\textbf{\begin{tabular}[c]{@{}c@{}}Detection\\ Success Rate (\%)\end{tabular}} & \multicolumn{1}{c|}{84.12} & \multicolumn{1}{c|}{80.44} & \multicolumn{1}{c|}{90.89} & \multicolumn{1}{c|}{89.99} & \multicolumn{1}{c|}{76.84} & \multicolumn{1}{c|}{71.68} & \multicolumn{1}{c|}{72.34} & \multicolumn{1}{c|}{82.98} & 74.44 \\ \hline
\textbf{\begin{tabular}[c]{@{}c@{}} $\mathbf{\zeta_{r}} $ (Px)\end{tabular}} & \multicolumn{1}{c|}{0.13} & \multicolumn{1}{c|}{0.19} & \multicolumn{1}{c|}{0.13} & \multicolumn{1}{c|}{0.17} & \multicolumn{1}{c|}{0.16} & \multicolumn{1}{c|}{0.21} & \multicolumn{1}{c|}{0.46} & \multicolumn{1}{c|}{0.51} & 0.43 \\ \hline
\end{tabular}%
}
\label{table:detection_rate}
\end{table*}

\begin{table*}[ht]
\centering
\caption{Calibration accuracy of of E-Calib eRWLS compared against OLS, MLESAC \cite{mlesac}, and RANSAC \cite{cylinder_ransac}, in terms of the reprojection error $\zeta_{r}$.}
\resizebox{\textwidth}{!}{%
\begin{tabular}{|c|cccccc|}
\hline
\multirow{2}{*}{\textbf{Method}} & \multicolumn{6}{c|}{\textit{\textbf{Dataset}}} \\ \cline{2-7} 
 & \multicolumn{1}{c|}{\textit{\textbf{\begin{tabular}[c]{@{}c@{}}davis\_3x7\_cone\\ gdlight\end{tabular}}}} & \multicolumn{1}{c|}{\textit{\textbf{\begin{tabular}[c]{@{}c@{}}davis\_3x7\_cone\\ lowlight\end{tabular}}}} & \multicolumn{1}{c|}{\textit{\textbf{\begin{tabular}[c]{@{}c@{}}davis\_3x9\_cone\\ gdlight\end{tabular}}}} & \multicolumn{1}{c|}{\textit{\textbf{\begin{tabular}[c]{@{}c@{}}davis\_3x9\_cone\\ lowlight\end{tabular}}}} & \multicolumn{1}{c|}{\textit{\textbf{\begin{tabular}[c]{@{}c@{}}davis\_4x11\_cone\\ gdlight\end{tabular}}}} & \textit{\textbf{\begin{tabular}[c]{@{}c@{}}davis\_4x11\_cone\\ lowlight\end{tabular}}} \\ \hline
\textbf{OLS} & \multicolumn{1}{c|}{0.39} & \multicolumn{1}{c|}{0.69} & \multicolumn{1}{c|}{0.40} & \multicolumn{1}{c|}{0.53} & \multicolumn{1}{c|}{0.43} & 0.49 \\ \hline
\textbf{MLESAC} & \multicolumn{1}{c|}{0.43} & \multicolumn{1}{c|}{0.48} & \multicolumn{1}{c|}{0.37} & \multicolumn{1}{c|}{0.36} & \multicolumn{1}{c|}{0.33} & 0.37 \\ \hline
\textbf{\begin{tabular}[c]{@{}c@{}}Hough Transform\\ with RANSAC\end{tabular}} & \multicolumn{1}{c|}{0.36} & \multicolumn{1}{c|}{0.37} & \multicolumn{1}{c|}{0.29} & \multicolumn{1}{c|}{0.31} & \multicolumn{1}{c|}{0.36} & 0.41 \\ \hline
\textbf{eRWLS} & \multicolumn{1}{c|}{0.13} & \multicolumn{1}{c|}{0.19} & \multicolumn{1}{c|}{0.13} & \multicolumn{1}{c|}{0.17} & \multicolumn{1}{c|}{0.16} & 0.21 \\ \hline
\end{tabular}%
}
\label{table:ols}
\end{table*}

\begin{figure*}[ht]
\center
\includegraphics[width=0.95\textwidth,height=0.55\textwidth,scale=1]{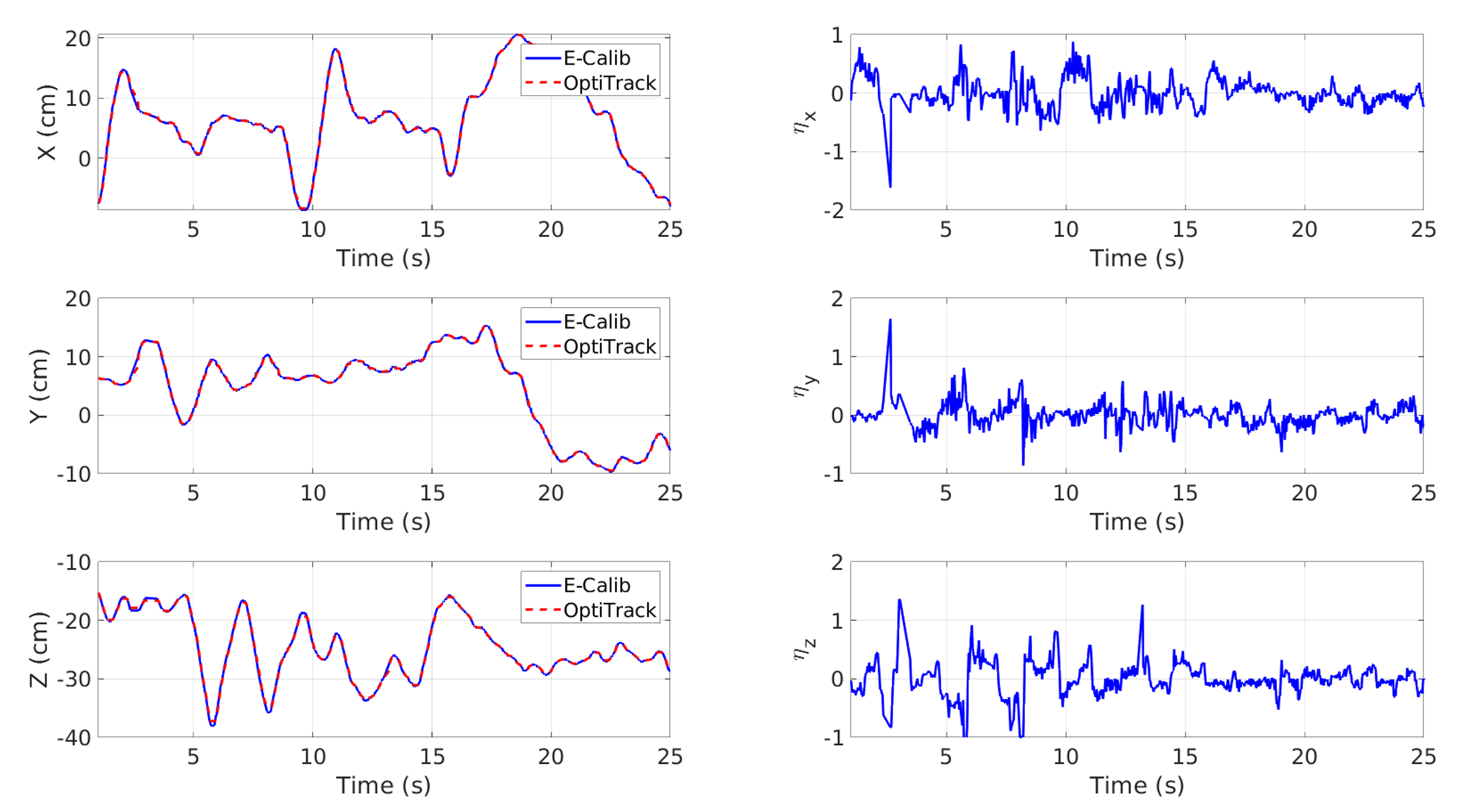}
\caption{Pose estimates by our method compared against ground truth pose of the camera from the motion capture system. Our approach achieves a mean absolute positioning error $\hat{\eta}_{t}$ of 0.952 cm and maximum positioning error $\eta^{max}_{t}$ of 1.72 cm.}
\label{fig:position_plot}
\end{figure*}

\subsection{Calibration Accuracy Evaluation} \label{subsection:accuracy}

To obtain optimal intrinsics of the event camera, 1) sufficient number of detections of the asymmetric circle grid is required and 2) the detected circle centers need to be localized with sub-pixel accuracy in the image plane. Accordingly, we compare our eRWLS feature extraction method to the detection performance of MATLAB's state-of-the-art detector, \textit{detectCircleGridPoints} \cite{find_circles}, applied on the frames of the DAVIS APS sensor and MR algorithm for \textit{davis\_4x11\_cone\_gdlight} and \textit{dvx\_4x11\_cone\_gdlight} datasets, demonstrated in Fig. \ref{fig:detection}. In addition, the detection success rate evaluates the performance of our modified hierarchical clustering algorithm to detect the circle grid even in the presence of scene background. \textit{detectCircleGridPoints} fails to detect the calibration pattern on the DAVIS frames due to the fact that motion blur is witnessed and the features of the calibration grid are degraded. On the other hand, even though the MR frames suffer from noise and artifacts significantly degrading the image quality and the calibration grid is not extracted.

To the contrary, our calibration tool, leveraging the asynchronous nature of events, detects the calibration pattern even when high speed motion is induced. We first evaluate the proposed calibration approach in terms of the detection success rate on \textit{ECam\_ACircles} dataset to ensure that calibration is attainable, where the detection success rate is defined as the fraction of times the pattern was detected against the total number of detections that can be obtained. It is worth noting that the total number of detections of the calibration pattern is equal to the data sequence time divided by the time step between the spatiotemporal windows, which we have defined as 33 ms. The detection success rates are shown in Table \ref{table:detection_rate} on the cone trajectory data sequences of the \textit{ECam\_ACircles} dataset. We also report the mean reprojection error for all DAVIS and DVXplorer data sequences in Table \ref{table:paths_repr}, where a mean reprojection error of 0.16 Px and 0.52 Px is obtained for DAVIS and DVXplorer calibrations, respectively. Notice that the detection success rate is high for both sensors showing that our approach proves to be resolution agnostic. In addition, the detection success rate is maintained even at challenging illumination conditions demonstrating that the proposed eRWLS is robust against noise and the modified hierarchical clustering algorithm identifies the calibration pattern despite the presence of the noisy background.

\begin{figure*}[ht]
\center
\includegraphics[width=0.95\textwidth,height=0.235\textheight,scale=1]{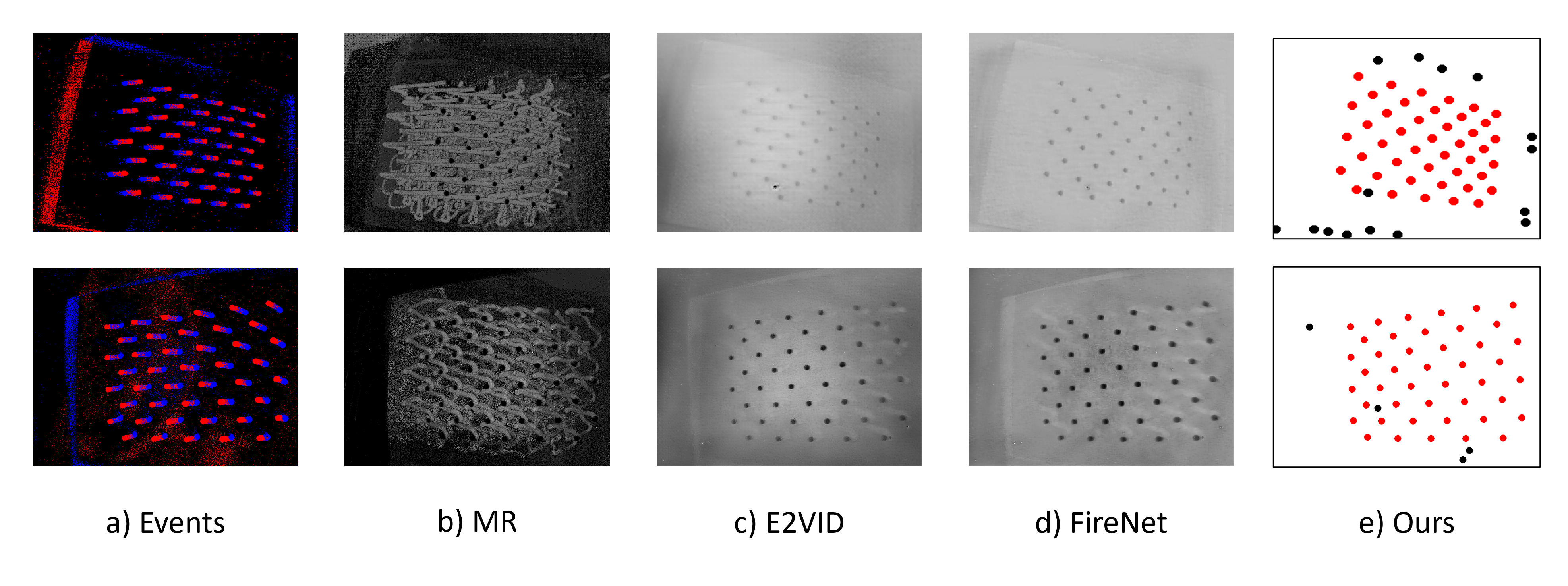}
\caption{Data sequences on the \textit{davis\_4x11\_cone\_gdlight} (top) and \textit{dvx\_4x11\_cone\_gdlight} (bottom), where state-of-the-art methods fail to detect the calibration pattern, while our method robustly detects the asymmetric circle grid.}
\label{fig:benchmarks}
\end{figure*}

\begin{table}[t]
\centering
\caption{The obtained intrinsics of the event camera by our method are compared against the parameters acquired by the DAVIS frames on {\textit{davis\_3x9\_cone\_gdlight}.}}
\resizebox{\columnwidth}{!}{%
\begin{tabular}{|c|c|c|}
\hline
\textbf{Metric} & \textbf{Frames} & \textbf{Ours} \\ \hline
$\zeta_{r}$ (Px) & 0.11 & 0.13 \\ \hline
\begin{tabular}[c]{@{}c@{}} $f_{x}, f_{y}$ (Px) \end{tabular} & 355.35, 354.31 & 355.54, 353.97 \\ \hline
\textit{\textbf{\begin{tabular}[c]{@{}c@{}}$u_{0},v_{0}$\end{tabular}}} & 159.84, 126.63 & 159.16, 124.45 \\ \hline
\textit{\textbf{$k_{1}$, $k_{2}$, $k_{3}$}} & -0.3469, 0.122, 0.1921 & -0.333, 0.075, 0.271 \\ \hline
\textit{\textbf{$p_{1}, p_{2}$}} & -0.000598, -0.000513 & -0.000670, -0.000596 \\ \hline
\textit{\begin{tabular}[c]{@{}c@{}}Detection\\ Success Rate (\%)\end{tabular}} & 79.92 & 84.12 \\ \hline
\end{tabular}%
}
\label{table:our_davis}
\end{table}

Having a sufficient number of pattern detections, the calibration accuracy is mainly dependent on the feature extraction performance of eRWLS since a state-of-the-art calibration optimization is utilized to obtain the sensor parameters. Hence, the calibration performance is evaluated in terms of the reprojection error and the deviation of the obtained intrinsic parameters from the conventional frames. The reprojection errors $\zeta_{r}$, evaluated similar to \cite{flexible_calib}, are reported in Table \ref{table:detection_rate} and Fig. \ref{fig:error}, while the obtained intrinsics compared against the conventional frames is reported in Table \ref{table:our_davis}. Note that this comparison is only applicable on \textit{davis\_3x9\_cone\_gdlight} because calibration was only successful using the DAVIS frames on the aforementioned dataset due to the slow sensor motion and adequate lighting conditions. The obtained results show that eRWLS extracts the pattern circle centers with the required accuracy. Moreover, low reprojection errors are maintained even in poorly illuminated environments, where sensor noise drastically increases.

We also motivate the use of eRWLS by comparing its performance to cylinder fitting using OLS, in terms of the reprojection error. In addition, we also provide comparisons with respect to state-of-the-art cylinder detection approaches, namely maximum-likelihood estimation by random sampling consensus (MLESAC) \cite{mlesac} and random sample consensus (RANSAC) \cite{cylinder_hough, cylinder_ransac} with Hough Transform, reported in Table \ref{table:ols} on the DAVIS cone path data sequence. Notice that OLS performs poorly in poor-lit conditions. On the other hand, MLESAC and RANSAC show consistent performance in poor in poorly lighted environments, still E-Calib outperforms these approaches in terms of the reprojection error $\zeta_{r}$, ensuring high calibration accuracy for event cameras.

\begin{table*}[!h]
\centering
\caption{Pose errors on cone trajectory data sequences of the \textit{ECam\_ACircles} dataset. Notice that high detection success rates and low reprojection errors are maintained even at challenging lighting conditions demonstrating robustness against noise, which is quite high in low illumination conditions.}
\resizebox{\textwidth}{!}{%
\begin{tabular}{|c|ccccccccc|}
\hline
\multirow{2}{*}{\textbf{Metric}} & \multicolumn{9}{c|}{\textit{\textbf{Dataset}}} \\ \cline{2-10} 
 & \multicolumn{1}{c|}{\textit{\textbf{\begin{tabular}[c]{@{}c@{}}davis\_3x7\\ gdlight\end{tabular}}}} & \multicolumn{1}{c|}{\textit{\textbf{\begin{tabular}[c]{@{}c@{}}davis\_3x7\\ lowlight\end{tabular}}}} & \multicolumn{1}{c|}{\textit{\textbf{\begin{tabular}[c]{@{}c@{}}davis\_3x9\\ gdlight\end{tabular}}}} & \multicolumn{1}{c|}{\textit{\textbf{\begin{tabular}[c]{@{}c@{}}davis\_3x9\\ lowlight\end{tabular}}}} & \multicolumn{1}{c|}{\textit{\textbf{\begin{tabular}[c]{@{}c@{}}davis\_4x11\\ gdlight\end{tabular}}}} & \multicolumn{1}{c|}{\textit{\textbf{\begin{tabular}[c]{@{}c@{}}davis\_4x11\\ lowlight\end{tabular}}}} & \multicolumn{1}{c|}{\textit{\textbf{\begin{tabular}[c]{@{}c@{}}dvx\_3x7\\ gdlight\end{tabular}}}} & \multicolumn{1}{c|}{\textit{\textbf{\begin{tabular}[c]{@{}c@{}}dvx\_3x9\\ gdlight\end{tabular}}}} & \textit{\textbf{\begin{tabular}[c]{@{}c@{}}dvx\_4x11\\ gdlight\end{tabular}}} \\ \hline
\textbf{\begin{tabular}[c]{@{}c@{}}$\mathbf{\hat{\eta}_{t}}$ (cm)\end{tabular}} & \multicolumn{1}{c|}{0.480} & \multicolumn{1}{c|}{1.190} & \multicolumn{1}{c|}{0.952} & \multicolumn{1}{c|}{0.721} & \multicolumn{1}{c|}{0.786} & \multicolumn{1}{c|}{0.947} & \multicolumn{1}{c|}{0.370} & \multicolumn{1}{c|}{1.048} & 0.510 \\ \hline
\textbf{\begin{tabular}[c]{@{}c@{}} $\mathbf{\sigma_{t}}$ (cm)\end{tabular}} & \multicolumn{1}{c|}{0.0319} & \multicolumn{1}{c|}{0.079} & \multicolumn{1}{c|}{0.083} & \multicolumn{1}{c|}{0.0751} & \multicolumn{1}{c|}{0.0506} & \multicolumn{1}{c|}{0.0686} & \multicolumn{1}{c|}{0.0851} & \multicolumn{1}{c|}{0.0459} & 0.0606 \\ \hline
\textbf{\begin{tabular}[c]{@{}c@{}}$\mathbf{\hat{\eta}_{r}}$ ($\degree$) \end{tabular}} & \multicolumn{1}{c|}{0.7730} & \multicolumn{1}{c|}{1.4964} & \multicolumn{1}{c|}{0.8290} & \multicolumn{1}{c|}{0.6970} & \multicolumn{1}{c|}{0.7362} & \multicolumn{1}{c|}{1.0491} & \multicolumn{1}{c|}{0.8981} & \multicolumn{1}{c|}{0.9367} & 0.3711 \\ \hline
\textbf{\begin{tabular}[c]{@{}c@{}} $\mathbf{\sigma_{r}}$ ($\degree$) \end{tabular}} & \multicolumn{1}{c|}{0.0347} & \multicolumn{1}{c|}{0.0436} & \multicolumn{1}{c|}{0.0435} & \multicolumn{1}{c|}{0.0312} & \multicolumn{1}{c|}{0.0284} & \multicolumn{1}{c|}{0.0504} & \multicolumn{1}{c|}{0.0573} & \multicolumn{1}{c|}{0.0497} & 0.0219 \\ \hline
\end{tabular}%
}
\label{table:position_error}
\end{table*}

Finally, as mentioned in section \ref{subsection:dbscan}, we have formulated spatiotemporal windows based on 4000 events. Even though though the number of events is set as a hyperparameter, we study the effect of the number of events on the calibration accuracy. Large number of events can lead to rather curved cylinders instead of slanted cylinders due to varying optical flow. Accordingly, we evaluate different number of events when formulating spatiotemporal windows and validate eRWLS feature extraction performance in terms of the reprojection error and detection success rate when curved cylinders are present, see Table \ref{table:events_ablation} reporting the calibration accuracy on \textit{dvx\_3x9\_cone\_gdlight}. Notice that the reprojection error is relatively consistent, even when large number of events, i.e. 15,000, leading to curved cylinders to be present. This demonstrates the robustness of eRWLS with the weighing factor that penalizes outliers, considered as points along the curved cylinder centerline.

\begin{table}[h]
\centering
\caption{Effect of the number of events on the eRWLS feature extraction performance on the \textit{dvx\_3x9\_cone\_gdlight} data sequence.}
\resizebox{0.35\textwidth}{!}{%
\begin{tabular}{|c|c|c|}
\hline
\textbf{No. Events} & \textit{$\zeta_{r}$ (Px)} & \textit{\begin{tabular}[c]{@{}c@{}}Detection\\ Success Rate (\%)\end{tabular}} \\ \hline
\textbf{4000} & 0.46 & 72.34 \\ \hline
\textbf{10000} & 0.49 & 70.21 \\ \hline
\textbf{15000} & 0.54 & 74.33 \\ \hline
\textbf{25000} & 0.47 & 77.21 \\ \hline
\end{tabular}%
}
\label{table:events_ablation}
\end{table}

\begin{table}[h]
\centering
\caption{Mean reprojection errors $\zeta_{r}$ obtained by E-Calib on all \textit{ECam\_ACircles} calibration trajectories.}
\resizebox{0.3\textwidth}{!}{%
\begin{tabular}{|c|c|c|}
\hline
\textbf{\begin{tabular}[c]{@{}c@{}}Data Sequence\\ Trajectory\end{tabular}} & \textit{\textbf{DAVIS}} & \textit{\textbf{DVXplorer}} \\ \hline
\textbf{Cone} & 0.14 & 0.47 \\ \hline
\textbf{Square} & 0.11 & 0.46 \\ \hline
\textbf{Spiral} & 0.18 & 0.54 \\ \hline
\textbf{\begin{tabular}[c]{@{}c@{}}Out-of-Focus\\ Scenes\end{tabular}} & 0.21 & 0.59 \\ \hline
\end{tabular}%
}
\label{table:paths_repr}
\end{table}

\begin{table}[t]
\centering
\caption{Mean translation $\hat{\eta}_{t}$ (cm) and rotation errors $\hat{\eta}_{r}$ $({}^{\circ})$ obtained by E-Calib on all \textit{ECam\_ACircles} calibration trajectories.}
\resizebox{0.4\textwidth}{!}{%
\begin{tabular}{|c|cl|cl|}
\hline
\multirow{2}{*}{\textbf{\begin{tabular}[c]{@{}c@{}}Data Sequence\\ Trajectory\end{tabular}}} & \multicolumn{2}{c|}{\textit{\textbf{DAVIS}}} & \multicolumn{2}{c|}{\textit{\textbf{DVXplorer}}} \\ \cline{2-5} 
 & \multicolumn{1}{l|}{$\hat{\eta}_{t}$ (cm)} & $\hat{\eta}_{r}$ (${}^{\circ}$) & \multicolumn{1}{l|}{$\hat{\eta}_{t}$ (cm)} & $\hat{\eta}_{r}$ (${}^{\circ}$) \\ \hline
\textbf{Cone} & \multicolumn{1}{c|}{0.916} & 0.930 & \multicolumn{1}{c|}{0.642} & 0.735 \\ \hline
\textbf{Square} & \multicolumn{1}{c|}{0.879} & 1.021 & \multicolumn{1}{c|}{0.777} & 0.822 \\ \hline
\textbf{Spiral} & \multicolumn{1}{c|}{0.992} & 0.991 & \multicolumn{1}{c|}{0.719} & 0.801 \\ \hline
\textbf{\begin{tabular}[c]{@{}c@{}}Out-of-Focus\\ Scenes\end{tabular}} & \multicolumn{1}{c|}{1.612} & 1.176 & \multicolumn{1}{c|}{1.113} & 0.939 \\ \hline
\end{tabular}%
}
\label{table:paths_pose}
\end{table}

\begin{table*}[ht]
\centering
\caption{The proposed event-based calibration algorithm is benchmarked against the works of Reinbacher et al \cite{event_mr}, Muglikar et al. \cite{eventcalib_e2vid}, Scheerlinck et al. \cite{event_firenet} and Huang et al. \cite{ecalib_iros} on the \textit{davis\_3x9\_cone\_gdlight}. The dash indicates the metric is not applicable while \Cross \text{} represents failure of obtaining the parameter.}
\resizebox{\textwidth}{!}{%
\begin{tabular}{|c|c|c|c|c|c|c|}
\hline
\textit{\textbf{Algorithm}} & \textit{\textbf{DAVIS Frames}} & \textit{\textbf{Reinbacher et al. \cite{event_mr}}} & \textit{\textbf{Muglikar et al. \cite{eventcalib_e2vid}}} & \textit{\textbf{Scheerlinck et al. \cite{event_firenet}}} & \textit{\textbf{Huang et al. \cite{ecalib_iros}}} & \textit{\textbf{Ours}} \\ \hline
Year & - & 2016 & 2019 & 2020 & 2021 & \textit{\textbf{2023}} \\ \hline
\begin{tabular}[c]{@{}c@{}}Reprojection\\ Error (Px) $\zeta_{r}$\end{tabular} & 0.11 & \Cross & 0.13 & 0.19 & 0.29 & \textit{\textbf{0.13}} \\ \hline
Focal Length ($f$) & 355.35 & \Cross & 372.28 & 360.71 & 357.091 & \textit{\textbf{355.54}} \\ \hline
$k_{1}$, $k_{2}$, $k_{3}$ & -0.344, 0.117, -0.0178 & \Cross & -0.081, 0.199, -0.313 & -0.291, 0.0367, -0.112 & -0.359, 0.396, -0.7178 & \textit{\textbf{-0.339, 0.071, -0.0271}} \\ \hline
$u_{0}$, $v_{0}$ & 159.84, 125.63 & \Cross & 164.23, 129.78 & 163.11, 128.24 & 158.98, 124.39 & \textit{\textbf{159.16, 124.45}} \\ \hline
Pattern Detections & 345 & 3 & 41 & 62 & - & \textit{\textbf{709}} \\ \hline
Detection Success Rate & 79.99\% & 0.38\% & 9.51\% & 13.91\% & - & \textit{\textbf{84.12}} \\ \hline
\end{tabular}%
}
\label{table:benchmarks}
\end{table*}

\subsection{Pose Estimation Results} \label{subsection:accuracy}

While in most cases the calibration reprojection error can serve as a good benchmark for our method, in some occasions calibration can converge to a local minimum with the reprojection error being relatively low but the obtained sensor parameters are not optimal. Thus, we also validate the proposed method in terms of the estimated extrinsics. The estimated camera pose by our method is compared to the sensor ground truth pose given by the OptiTrack motion capture system in terms of the mean absolute error. The results are reported in Table \ref{table:position_error} on \textit{ECam\_ACircles} dataset in terms of the mean absolute translation $\hat{\eta}_{t}$ and rotation errors $\hat{\eta}_{r}$, and their corresponding standard deviations $\sigma_{t}$ and $\sigma_{r}$. Fig. \ref{fig:position_plot} also shows the positioning plots compared against the OptiTrack for \textit{davis\_3x9\_cone\_gdlight}. In addition, the mean pose error on all DAVIS and DVXplorer data sequences for all calibration trajectories is reported in table \ref{table:paths_pose}. Accurate pose estimation is directly correlated with robust tracking of the calibration targets centers. If ST-DBSCAN cluster centers are utilized as circle centers, $\zeta_{r}$ grows to 0.44 pixels and $\hat{\eta}_{t}$ rises to 6.71 cm on \textit{davis\_3x9\_cone\_gdlight}. This demonstrates the need for eRWLS refinement since ST-DBSCAN cluster centers are affected by severe sensor noise, especially in challenging lighting conditions. Moreover, OLS, MLESAC \cite{mlesac}, and Hough Transform with RANSAC \cite{cylinder_ransac} provide mean translation errors of 2.431 cm, 1.983, 1.912 cm on DAVIS data sequences, respectively, while our method provides a mean translation error of 0.973 cm. Our proposed method provides accurate estimates of the sensor pose and maintains a pose estimation error in poorly lighted environments of the same order of magnitude compared to data sequences collected at good lighting conditions. This shows robustness against the event camera noise, which increases with decreasing illumination.

\subsection{Benchmarks} \label{subsection:benchmarks}
To evaluate the impact of our event-driven calibration tool, we compare our results to state-of-the-art methods that include E2VID network by Muglikar et al. \cite{eventcalib_e2vid}, FireNet by Scheerlinck et al. \cite{event_firenet}, and the MR approach by Reinbacher et al. \cite{event_mr}. The methods are compared to the work of this paper in terms of the detection success rate, reprojection error, and pose error if calibration was successful by a designated method. Fig. \ref{fig:benchmarks} compares the detection performance of our method against the aforementioned works. An important point is that detection success rate was not evaluated on the work of Huang et al. \cite{ecalib_iros} since the required number of pattern detections is preset before running calibration. We have set this value to 700 sufficient to perform proper calibration, to compare the obtained results fairly and such that the total number of pattern detections are within the same order of magnitude. Table \ref{table:benchmarks} lists our results compared to those works on the \textit{davis\_3x9\_cone\_gdlight} dataset. In addition, only the radial distortion were computed for the benchmarks, while tangential distortion was ignored for all methods since it is not computed by Huang et al. \cite{ecalib_iros}. It is also worth mentioning that these approaches were tested on the \textit{davis\_3x9\_cone\_gdlight} by the codes publicly available by the authors of these papers.

\begin{figure}[t]
\center
\includegraphics[scale=0.27]{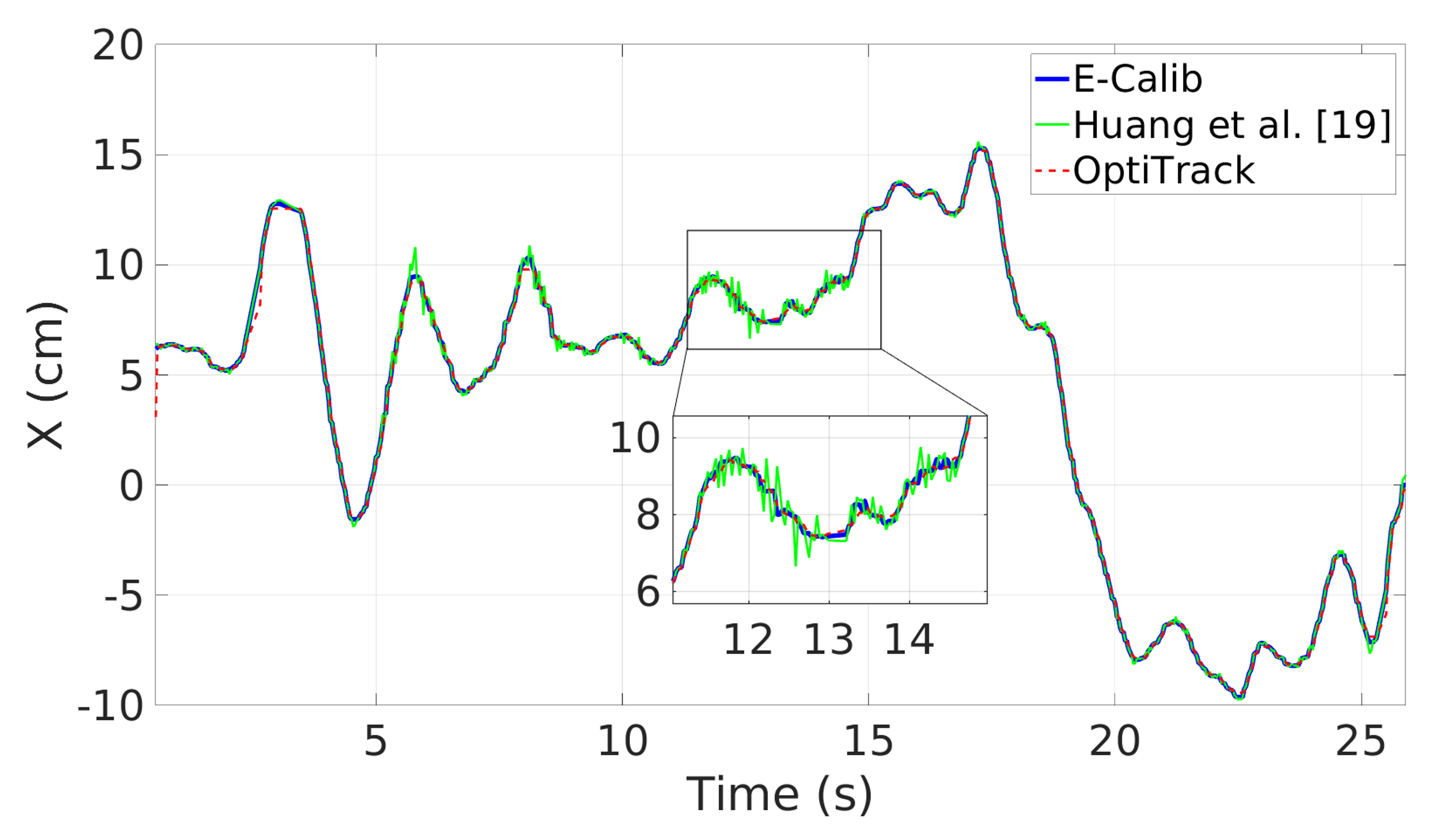}
\caption{Camera pose estimate plot along the Y-axis of our method, the work of Huang et al. \cite{ecalib_iros}, and ground truth from OptiTrack. Notice the pose estimates are smoother due to our improved method of tracking the circular features.}
\label{fig:pose_calib}
\end{figure}

Our approach outperforms the state-of-the-art significantly in terms of reprojection error, detection success rate, and intrinsic parameters values. The manifold regularization method by Muglikar et al. \cite{event_mr} failed to calibrate the camera and detect the calibration grid due to artifacts and sensor noise present in the reconstructed images. On the other hand, calibration was attained using E2VID \cite{eventcalib_e2vid} and FireNet \cite{event_firenet}, but the intrinsics deviate away from the parameters obtained by the frames as the calibration converged to a local minimum, even though a low reprojection error is obtained. This occured because most of the pattern detections were redundant and the calibration grid was not sufficiently detected from multiple views due to the degraded features of the circles in the reconstructed images, see Fig. \ref{fig:benchmarks}. Note that MATLAB's \textit{detectCircleGridPoints} \cite{find_circles} was utilized to extract the calibration grid from the reconstructed images. Finally, calibration was successful by Huang et al. \cite{ecalib_iros} method. Yet, our approach provides improved calibration accuracy also in terms of reprojection error and the intrinsics. Even though the obtained focal length by Huang's method is close to the focal length obtained by the DAVIS frames, the radial distortion coefficients deviate from their true values. This is because the aforementioned appraoch relies on the raw events as control points, where sub-pixel localization of the circle centers is not attained leading to sub-optimal calibration accuracy.

In addition to the intrinsic parameters, the obtained extrinsics were also validated against the ground truth pose of the camera and compared to the camera pose estimation results by our method, Huang's method, E2VID, and FireNet. Fig. \ref{fig:pose_calib} shows the positioning plot on \textit{davis\_3x9\_gdlight} along the y-axis and Table \ref{table:position_eventcalib} quantifies these results with the same metrics reported in Table \ref{table:position_error}. Table \ref{table:position_eventcalib} also reports the positioning results of E2VID and FireNet, where a large positioning error was obtained due to the sub-optimal intrinsic parameters. On the other hand, Huang's method and our approach provide satisfactory results in terms of a mean positioning error. However, our calibration tool shows a smoother estimate of camera trajectory, demonstrating the capability of our method to provide more optimal sensor intrinsics. This occured due to the sub-optimal optimization of the lens distortion coefficients and illustrates that sub-pixel localization accuracy of the circle centers is necessary for optimal sensor calibration, which is attained by our method unlike Huang's approach.

\begin{table}[h]
\centering
\caption{Mean absolute pose error by our method and the works of Muglikar et al. \cite{eventcalib_e2vid}, Scheerlinck et al. \cite{event_firenet}, and Huang et al. \cite{ecalib_iros} compared against ground truth camera pose obtained from OptiTrack.}
\resizebox{\columnwidth}{!}{%
\begin{tabular}{|c|c|c|c|c|}
\hline
\textbf{Method} & \textbf{Huang et al.} & \textbf{Muglikar et al.} & \textbf{Scheerlinck et al.} & \textbf{Ours} \\ \hline
\textit{\begin{tabular}[c]{@{}c@{}}$\hat{\eta}_{t}$ (cm)\end{tabular}} & 1.97 & 4.49 & 3.95 & \textbf{0.952} \\ \hline
\textit{\begin{tabular}[c]{@{}c@{}}$\hat{\sigma}_{t}$ (cm)\end{tabular}} & 0.129 & 0.396 & 0.218 & \textbf{0.083} \\ \hline
\textit{\begin{tabular}[c]{@{}c@{}}$\eta_{t}^{max}$ (cm)\end{tabular}} & 3.18 & 8.14 & 7.69 & \textbf{1.72} \\ \hline
\textit{\begin{tabular}[c]{@{}c@{}}$\hat{\eta}_{r}$ ($\degree$)\end{tabular}} & 2.31 & 5.37 & 4.99 & \textbf{0.8290} \\ \hline
\textit{\begin{tabular}[c]{@{}c@{}}$\hat{\sigma}_{r}$ ($\degree$)\end{tabular}} & 0.0913 & 0.271 & 0.179 & \textbf{0.0435} \\ \hline
\textit{\begin{tabular}[c]{@{}c@{}}$\eta_{r}^{max}$ ($\degree$)\end{tabular}} & 3.17 & 9.77 & 8.56 & \textbf{1.13} \\ \hline
\end{tabular}%
}
\label{table:position_eventcalib}
\end{table}

The average execution time of the proposed work for datasets on the DAVIS346 and DVXplorer is also reported for each algorithm block, see Table \ref{table:exec_time}. Compared to E2VID and FireNet networks, which run offline and require GPU acceleration, our algorithm outperforms these approaches in execution time without the need for heavy computational resources. On the other hand, while the MR approach achieves real-time performance, it suffers in terms of feature extraction performance and detection success rate. Nevertheless, our approach can surely be utilized for online calibration of the sensor. In addition, real-time performance is unnecessary for intrinsic calibration as redundant frames can be captured in close time steps. Nevertheless, it is worth mentioning that our algorithm is written in Python and MATLAB and was executed on Ryzen 7 6800H (3.2 GHz) processor. Thus, our execution time can be further improved if implemented using C++, especially if execution in real-time is needed. In addition, since we employ an exponential fitting function in eRWLS, the execution time can be further reduced by employing a polynomial approximation instead of the exponential function.

\begin{table}[h]
\centering
\caption{Execution times for the system blocks.}
\resizebox{0.75\columnwidth}{!}{%
\begin{tabular}{|c|c|c|}
\hline
\textit{\textbf{Algorithm}}                                           & \textit{\textbf{DAVIS346 (s)}} & \textit{\textbf{DVXplorer (s)}} \\ \hline
ST-DBSCAN                                                                & \textit{0.043}                      & \textit{0.067}                       \\ \hline
\begin{tabular}[c]{@{}c@{}}Efficient Reweighted\\ Least Squares\end{tabular} & \textit{0.016}                      & \textit{0.029}                       \\ \hline
\begin{tabular}[c]{@{}c@{}}Modified Hierarchical \\ Clustering\end{tabular}     & \textit{0.0012}                     & \textit{0.0014}                      \\ \hline
\textbf{Total}                                                        & \textbf{0.0602}                     & \textbf{0.0974}                      \\ \hline
\end{tabular}
}
\label{table:exec_time}
\end{table}

\section{Conclusion} \label{section:conc}
In this paper, a robust calibration tool is proposed for obtaining the intrinsic parameters of event cameras. Without relying on conventional image processing techniques, the demonstrated approach leverages the asynchronous nature and low latency of the event camera to provide robust estimates of the sensor intrinsics. This was attained by a resolution agnostic density-based spatiotemporal clustering algorithm. More importantly, a novel and efficient reweighted least squares (eRWLS) method was proposed to extract the features of the calibration targets despite the presence of sensor noise. The robustness and ability of eRWLS to track the circular features with sub-pixel accuracy played an important role in acquiring the desired calibration accuracy and outperforming state-of-the-art methods for event cameras calibration. In addition, our calibration tool introduced a modified hierarchical clustering algorithm to detect the calibration grid even in the presence of background clutter. The proposed method was tested in a variety of rigorous experiments on different sensor models, circle grids of different geometric properties, and under challenging illumination environments. The experiments demonstrated that our method outperformed the state of the art in terms of detection success rate, reprojection, and pose error. While the proposed work was utilized for calibrating event cameras, it can be utilized for event-based circle detection, a fundamental exigency in robot perception algorithms. Future work will also extend this work for calibrating stereo event cameras and extrinsic calibration of event cameras in robotic systems.

\bibliographystyle{IEEEtran}
\bibliography{main.bib}

\vskip -1\baselineskip plus -1fil

\begin{IEEEbiography}[{\includegraphics[width=1in,height=1.25in,clip,keepaspectratio]{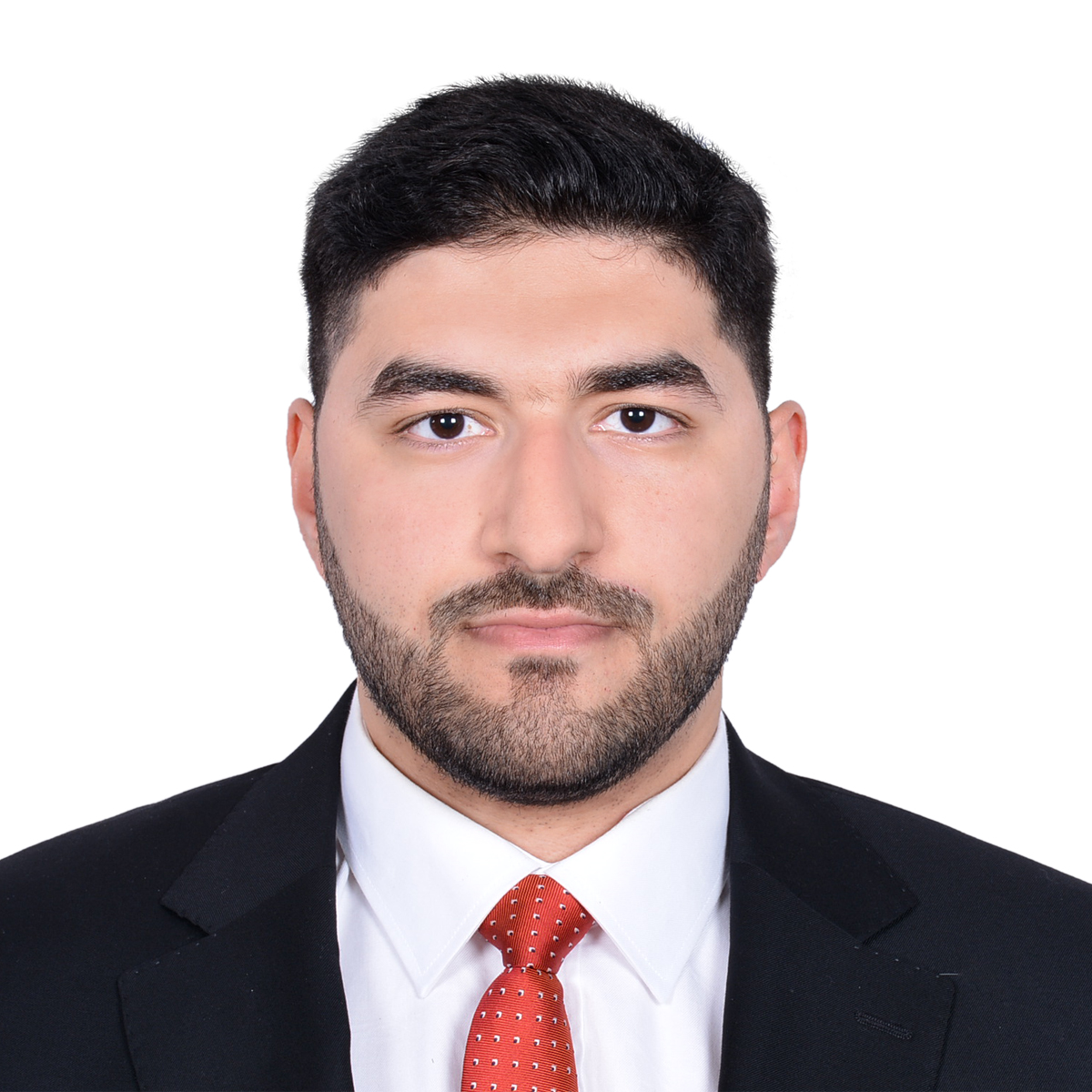}}]%
{Mohammed Salah}
received his BSc. in Mechanical Engineering from American University of Sharjah, UAE, in 2020 and his MSc. in Mechanical Engineering from Khalifa University, Abu Dhabi, UAE, in 2022. He is currently with Khalifa University Center for Autonomous Robotic Systems (KUCARS). His research interest is mainly focused on multi-sensor fusion, neuromorphic vision, and space robotics.
\end{IEEEbiography}

\vskip -1\baselineskip plus -1fil

\begin{IEEEbiography}[{\includegraphics[width=1in,height=1.25in,clip,keepaspectratio]{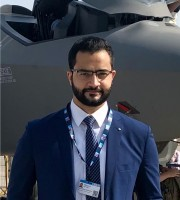}}]%
{Abdulla Ayyad}
(Member, IEEE) received the M.Sc. degree in electrical engineering from The University of Tokyo, in 2019, where he conducted research with the Spacecraft Control and Robotics Laboratory. He is currently a Research Associate at the Advanced Research and Innovation Center (ARIC) at Khalifa University working on several robot autonomy projects. His current research interest includes the application of AI in the fields of perception, navigation, and control.
\end{IEEEbiography}

\begin{IEEEbiography}[{\includegraphics[width=1in,height=1.25in,clip,keepaspectratio]{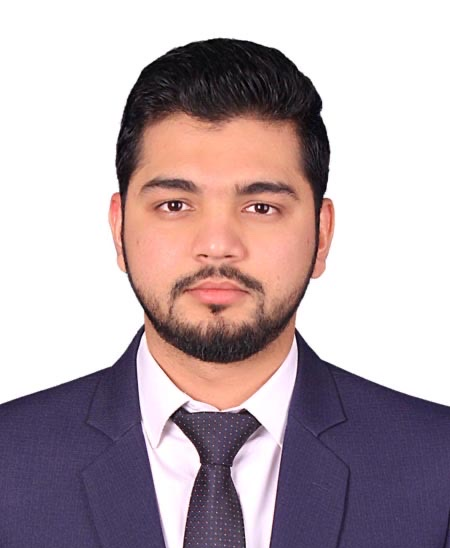}}]%
{Muhammad Humais}
received his M.Sc. in Electrical and Computer Engineering from Khalifa University in 2020. His research is mainly focused on robotic perception and control for autonomous systems. He is currently a Ph.D. fellow at Khalifa University Center for Autonomous Robotics (KUCARS). 
\end{IEEEbiography}

\begin{IEEEbiography}[{\includegraphics[width=1in,height=1.25in,clip,keepaspectratio]{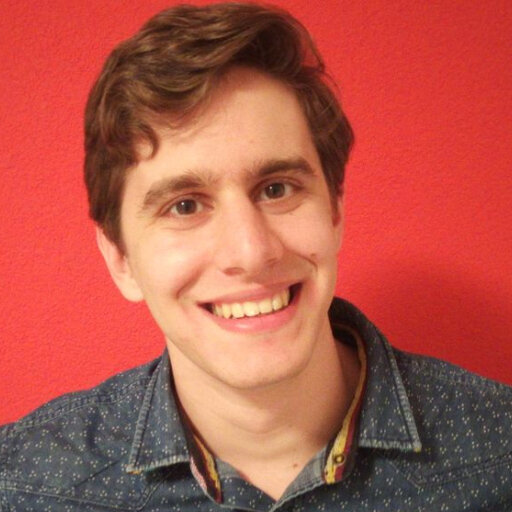}}]%
{Daniel Gehrig} received his BSc. and MSc. in Mechanical Engineering from the ETH Zurich, Switzerland, in 2019. He is currently with the University of Zurich at the Robotics and Perception Group. His research interests are focused on exploring the intersection of event-based vision, deep learning, and robotics.
\end{IEEEbiography}

\begin{IEEEbiography}[{\includegraphics[width=1in,height=1.25in,clip,keepaspectratio]{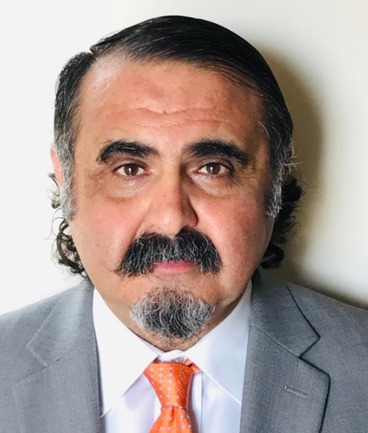}}]%
{Abdelqader Abusafieh} is currently SVP for Technology \& Advanced Materials at Strata Manufacturing PJSC (a Mubadala Company) where he is responsible for driving R\&D strategy and technology development programs within Mubadala Aerospace assets including collaboration initiatives with OEMs, technology partners, and academia. Prior to this role, he worked as a Technical Advisor at Mubadala Aerospace \& Defense unit for investment activities and Technology and Training initiatives. 
Dr. Abusafieh received his Master’s in Mechanical Engineering from Villanova University his Ph.D. in Materials Engineering from Drexel University. He has several patents and numerous publications and invited seminars. He sits on a number of senior management boards in academia and industry. 
\end{IEEEbiography}

\begin{IEEEbiography}[{\includegraphics[width=1in,height=1.25in,clip,keepaspectratio]{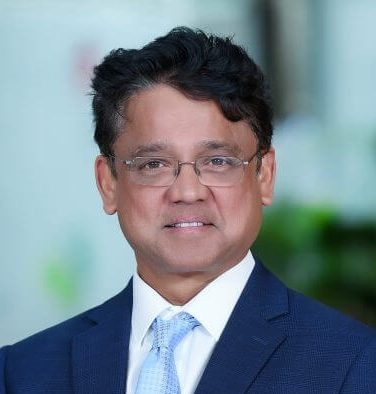}}]%
{Lakmal Seneviratne}
is Professor of Mechanical Engineering and the founding Director of the Centre for Autonomous Robotic Systems (KUCARS) at Khalifa University, UAE. He has also served as Associate Provost for Research and Graduate Studies and Associate VP Research at Khalifa University. Prior to joining Khalifa University, he was Professor of Mechatronics, the founding Director of the Centre for Robotics Research and the Head of the Division of Engineering, at King’s College London. He is Professor Emeritus at King’s College London. His main research interests are centered on robotics and automation, with particular emphasis on increasing the autonomy of robotic systems interacting with complex dynamic environments. He has published over 400 peer reviewed publications on these topics. He is a member of the Mohammed Bin Rashid Academy of Scientists in the UAE.
\end{IEEEbiography}

\vskip -1\baselineskip plus -1fil

\begin{IEEEbiography}[{\includegraphics[width=1in,height=1.25in,clip,keepaspectratio]{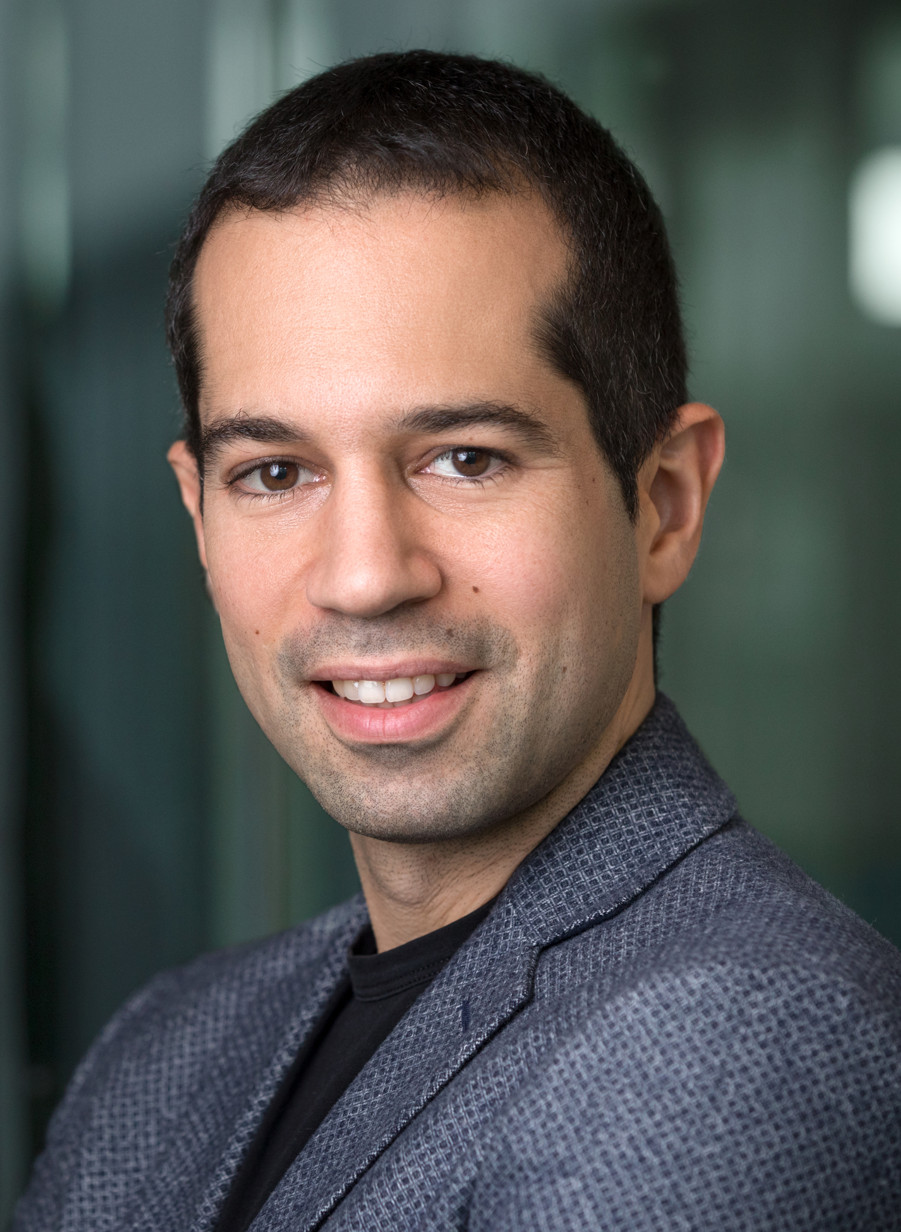}}]%
{Davide Scaramuzza} (1980, Italy) received a Ph.D. degree in robotics and computer vision from ETH Zurich, Switzerland, in 2008, followed by postdoctoral research at both ETH Zurich and the University of Pennsylvania, Philadelphia, USA. He is a Professor of Robotics and Perception with the University of Zurich, where he does research at the intersection of robotics, computer vision, and machine learning, aiming to enable autonomous, agile navigation of micro drones using standard and neuromorphic event-based cameras. From 2009 to 2012, he led the European project sFly, achieving the first autonomous vision-based navigation of microdrones in GPS-denied environments, which inspired the visual-navigation algorithm of the NASA Mars helicopter. He has served as a consultant for the United Nations on topics such as disaster response and disarmament, as well as the Fukushima Action Plan on Nuclear Safety. He coauthored the book Introduction to Autonomous Mobile Robots (MIT Press). For his research contributions, he won prestigious awards, such as a European Research Council (ERC) Consolidator Grant, the IEEE Robotics and Automation Society Early Career Award, an SNF-ERC Starting Grant, a Google Research Award, a Facebook Distinguished Faculty Research Award, and several paper awards. In 2015, he co-founded Zurich-Eye, today Facebook Zurich, which developed the hardware and software tracking modules of the
Oculus VR headset, which sold over 10 million units. Many aspects of his research have been prominently featured in wider media, such as The New York Times, The Economist, Forbes, BBC News, Discovery Channel.
\end{IEEEbiography}

\vskip -1\baselineskip plus -1fil

\begin{IEEEbiography}[{\includegraphics[width=1in,height=1.25in,clip,keepaspectratio]{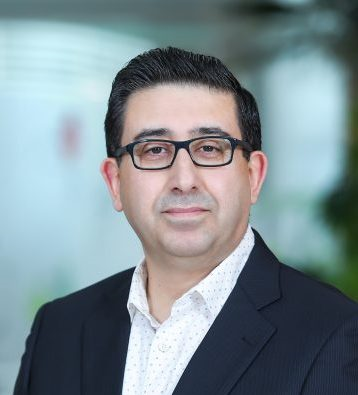}}]%
{Yahya Zweiri} (Member, IEEE), obtained his Ph.D. degree from King's College London. He is currently a Professor at the Department of Aerospace Engineering and the Director of the Advanced Research and Innovation Center, Khalifa University, United Arab Emirates. Over the past two decades, he has actively participated in defense and security research projects at institutions such as the Defense Science and Technology Laboratory, King's College London, and the King Abdullah II Design and Development Bureau in Jordan. Dr. Zweiri has a prolific publication record, with over 130 refereed journals and conference papers, as well as ten filed patents in the USA and UK. His primary research focus centers around robotic systems for challenging environments, with a specific emphasis on applied AI and neuromorphic vision systems.
\end{IEEEbiography}

\end{document}